\documentclass[10pt,journal]{IEEEtran}
\usepackage{amsmath,amsfonts}
\usepackage{algorithmic}
\usepackage{algorithm}
\usepackage{array}
\usepackage[caption=false,font=normalsize,labelfont=sf,textfont=sf]{subfig}
\usepackage{textcomp}
\usepackage{stfloats}
\usepackage{url}
\usepackage{verbatim}
\usepackage{graphicx}
\usepackage{cite}
\usepackage{amssymb}

\usepackage{multirow}
\usepackage{bbding}
\usepackage{tabularx}
\usepackage{float}
\usepackage{color}
\usepackage{tikz,xcolor,hyperref}
\definecolor{lime}{HTML}{A6CE39}
\DeclareRobustCommand{\orcidicon}{%
    \begin{tikzpicture}
    \draw[lime, fill=lime] (0,0) 
    circle [radius=0.16] 
    node[white] {{\fontfamily{qag}\selectfont \tiny ID}};    \draw[white, fill=white] (-0.0625,0.095) 
    circle [radius=0.007];    \end{tikzpicture}
    \hspace{-2mm}}
\foreach \x in {A, ..., Z}{%
    \expandafter\xdef\csname orcid\x\endcsname{\noexpand\href{https://orcid.org/\csname orcidauthor\x\endcsname}{\noexpand\orcidicon}}
    }

\begin{document}

\title{Learning Constrained Dynamic Correlations in Spatiotemporal Graphs for Motion Prediction}

\author{Jiajun Fu\orcidA{}, Fuxing Yang\orcidB{}, Yonghao Dang\orcidC{}, Xiaoli Liu\orcidD{}, and Jianqin Yin$^{*}$\thanks{* Jianqin Yin is the corresponding author.}\orcidE{}
\thanks{This work was supported partly by the National Natural Science Foundation of China (Grant No. 62173045, 62273054), partly by the Fundamental Research Funds for the Central Universities (Grant No. 2020XD-A04-3), and the Natural Science Foundation of Hainan Province (Grant No. 622RC675).}
\thanks{All authors are with the School of Artificial Intelligence, Beijing University of Posts and Telecommunications, Beijing, China.}
\thanks{This article has supplementary material provided by the authors and color versions of one or more figures available at https://doi.org/10.1109/TNNLS.2023.3277476}
\thanks{Digital Object Identifier 10.1109/TNNLS.2023.3277476}
}

{}
\markboth{IEEE Transactions on Neural Network and Learning System}
{Fu \MakeLowercase{\textit{et al.}}: Dynamic SpatioTemporal Decompose Graph Convolution Network.}


\maketitle

\begin{abstract}
  Human motion prediction is challenging due to the complex spatiotemporal feature modeling. Among all methods, Graph Convolution Networks (GCNs) are extensively utilized because of their superiority in explicit connection modeling. Within a GCN, the graph correlation adjacency matrix drives feature aggregation and thus is the key to extracting predictive motion features. State-of-the-art methods decompose the spatiotemporal correlation into spatial correlations for each frame and temporal correlations for each joint. Directly parameterizing these correlations introduces redundant parameters to represent common relations shared by all frames and all joints. Besides, the spatiotemporal graph adjacency matrix is the same for different motion samples and thus cannot reflect sample-wise correspondence variances. To overcome these two bottlenecks, we propose Dynamic SpatioTemporal Decompose Graph Convolution (DSTD-GC), which only takes 28.6\% parameters of the state-of-the-art graph convolution. The key of DSTD-GC is constrained dynamic correlation modeling, which explicitly parameterizes the common static constraints as a spatial/temporal vanilla adjacency matrix shared by all frames/joints and dynamically extracts correspondence variances for each frame/joint with an adjustment modeling function. For each sample, the common constrained adjacency matrices are fixed to represent generic motion patterns, while the extracted variances complete the matrices with specific pattern adjustments. Meanwhile, we mathematically reformulate graph convolutions on spatiotemporal graphs into a unified form and find that DSTD-GC relaxes certain constraints of other graph convolutions, which contributes to a better representation capability. Moreover, by combining DSTD-GC with prior knowledge like body connection and temporal context, we propose a powerful spatiotemporal graph convolution network called DSTD-GCN. On the Human3.6M, CMU Mocap, and 3DPW datasets, DSTD-GCN outperforms state-of-the-art methods by 3.9\% - 8.7\% in prediction accuracy with 55.0\% - 96.9\% fewer parameters. Codes are available at https://github.com/Jaakk0F/DSTD-GCN.
\end{abstract}


\begin{figure}[htbp]
  \centering
  \includegraphics[width=0.48\textwidth]{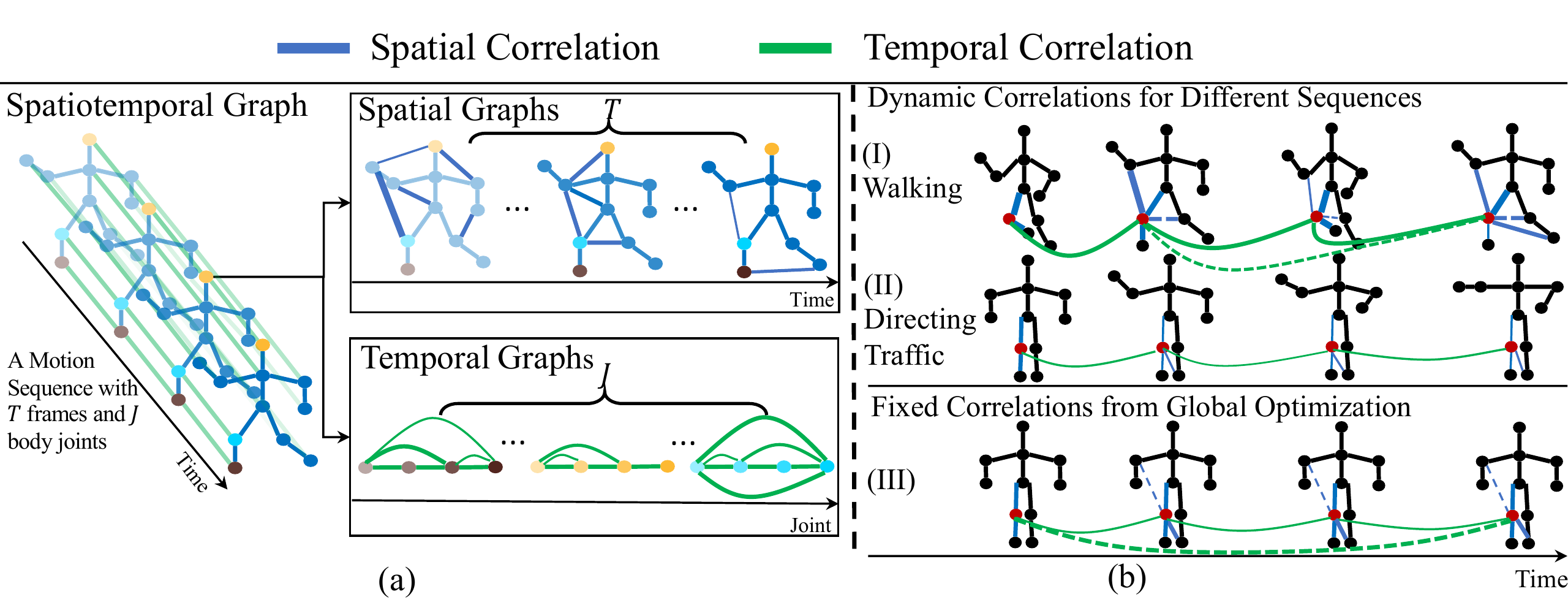}
  \vspace{-0.4cm}
  \caption{(a) Spatiotemporal-unshared decomposition. A spatiotemporal motion graph with $T$ frames and $J$ joints can be decomposed into $J$ temporal graphs and $T$ spatial graphs. The spatial/temporal correlations are different from each frame/joint. The skeleton color brightness in spatial graphs indicates the corresponding frame and the joint color in temporal graphs indicates the corresponding joint. (b) Dynamic correlation modeling. We use the correlation for the red knee joint as an example. Correlations vary for different actions, and the correlations in walking (I) are different from those in directing traffic (II). These variations cannot be explicitly captured by the learned correlations from global optimization (III), thus introducing correlation amplification or reduction (The dashed line).}
  \label{fig:motivation}
  \vspace{-0.8cm}
\end{figure}

\section{Introduction}
  
\IEEEPARstart{H}{uman} motion prediction is a challenging task that involves modeling dynamic spatiotemporal correlations between body joints. This task is becoming increasingly important in autonomous driving \cite{paden2016survey}, human-machine interaction \cite{kong2018human,unhelkar2018human, yan2022probabilistic}, and healthcare \cite{troje2002decomposing}.
  
Traditional human motion prediction methods can handle simple and periodic scenarios with classical time series processing approaches \cite{taylor2010dynamical,lehrmann2014efficient}, but these methods make unrealistic predictions when human motion becomes complex and erratic. Recently, researchers have adopted different deep-learning techniques for human motion prediction, such as Convolution Neural Network (CNN) \cite{li2018convolutional,liu2020trajectorycnn}, Recurrent Neural Network (RNN) \cite{al2020attention,guo2019human,martinez2017human,fragkiadaki2015recurrent,martinez2017human,li2018convolutional,tang2018long,liu2019towards,pavllo2020modeling,shu2021spatiotemporal,wang2021pvred}, Generative Adversarial Network (GAN) \cite{gui2018adversarial,hernandez2019human,cui2021efficient,ke2019learning,kundu2019bihmp} and Transformer \cite{cai2020learning,aksan2021spatio,tevet2023human}. Though these methods have achieved significant advancements over traditional approaches, they cannot explicitly represent connections among body joints and thus fail to model inherent body joint relations effectively.

With joints as vertices, bone and trajectory connections as edges, a human motion sequence can be naturally represented as a spatiotemporal graph. Graph Convolution Networks (GCNs) have been applied to motion prediction \cite{mao2019learning,mao2020history,li2020dynamic,cui2020learning,dang2021msr,liu2021motion,sofianos2021space,li2022online,dong2022skeleton}. Within the GCN, the spatiotemporal correlation guides the feature aggregation and thus is the key to achieving high predictability. For the spatiotemporal correlation, flexible and various motion sequences are represented by two crucial characteristics. The first one is \emph{spatiotemporal-unshared decomposition} (Fig \ref{fig:motivation} (a)). Since a joint is physically connected to its kinetic neighbors and virtually connected to its trajectory neighbors\cite{yan2018spatial,sofianos2021space},
we can decompose a spatiotemporal motion graph into several spatial and temporal graphs. The spatial graphs model intra-body relations for each frame, and the temporal graphs model inter-frame relations for each joint.
With this decomposition, spatiotemporal feature modeling is conducted by alternatively updating features on spatial and temporal graphs. This strategy reduces the complexity of spatiotemporal graph adjacency representation and has been widely adopted. For the spatiotemporal-unshared decomposition, the spatial graph correlations are varied at each frame to reflect changing motion patterns, while the temporal graph correlations are different at each joint to depict distinct trajectory patterns. We refer to these correlations as unshared spatial and temporal correlations. This precise representation contributes to predictive feature extraction.
The second characteristic is \emph{sample-specific representation}, where spatiotemporal graph adjacency is adjusted for different motion sequences. Since the motion patterns are varied for different motion sequences, the spatiotemporal correlation should precisely reflect sample-specific patterns in the spatiotemporal graph.

For spatiotemporal-unshared decomposition, Sofianos \textit{et al.} \cite{sofianos2021space} proposed the only spatiotemporal-unshared method, where a cross-talk between unshared spatial and temporal correlations depict the connections in a spatiotemporal graph. This method significantly improves the representation ability of Graph Convolutions (GCs) and outperforms previous methods with far fewer parameters. However, it still introduces redundant parameters for representing common prior knowledge, such as body connections and temporal context. \emph{There is still room to explore more parameter-saving representations of spatiotemporal correlations.} For sample-specific representations, despite the plausible motive behind this design, previous works all \cite{mao2019learning,mao2020history,li2020dynamic,cui2020learning,liu2021motion,dang2021msr,sofianos2021space} adopted generic graph correlations across input sequences. These sample-generic representations are learned through back-propagation and are optimal for all data, but they may disturb individual predictions by mistakenly enhancing unrelated correspondences or reducing critical connection strengths (Fig. \ref{fig:motivation} (b)). The correspondence disturbance amplifies the slight movement or the inability to infer future motion from tiny cues. Therefore, \emph{it is urgent to develop methods with sample-specific spatiotemporal correspondence.}

To overcome the aforementioned challenges, we look into human motions' \emph{constrained dynamic relational nature}, a critical characteristic ignored by all previous GCN-based methods. For every sample, the relations between body joints are subject to common static constraints such as body connections and temporal context (e.g., a frame connects to it adjacent one). Meanwhile, the relation strengths between certain joints are dynamically enhanced or reduced according to different samples.

Inspired by the above discussions, we introduce Dynamic SpatioTemporal Decompose Graph Convolution (DSTD-GC), which contains Dynamic Spatial Graph Convolution (DS-GC) and Dynamic Temporal Graph Convolution (DT-GC) for constrained dynamic spatial and temporal correlation modeling. The DS-GC/DT-GC is equipped with constrained dynamic correlation modeling, which decouples correlations into static constraints and dynamic adjustments. The static constrained correlation is parameterized as a vanilla spatial/temporal adjacency matrix, and it is shared by all frames/joints across all samples. This sample-generic shared correlation serves as motion prior and indicates general relations between vertices, and it is completed by sample-specific unshared correspondence adjustments. These dynamic adjustments are extracted by an adjustment modeling function based on the input sample. According to this formulation, we represent spatiotemporal-unshared correlations in quarter space complexity, while the previous representation \cite{sofianos2021space} is of cubic space complexity. Besides, DSTD-GC captures dynamic motion patterns between different samples and adjusts correlations between graph vertices, which precisely enhances correspondences for mutual motions and reduces correspondences for unrelated motions. Furthermore, we analyze different types of spatiotemporal graph convolutions in a unified form and theoretically prove that DSTD-GC enhances feature representation by relaxing certain restrictions of spatiotemporal graph convolutions. Finally, by combining DSTD-GC and prior knowledge, such as body connections and temporal context, we propose DSTD-GCN, which outperforms other state-of-the-art methods in prediction accuracy with the fewest parameters.
  
The contributions of this paper are summarized as follows: 
\begin{itemize}
    \item Dynamic SpatioTemporal Decompose Graph Convolution (DSTD-GC) is proposed, which utilizes constrained dynamic correlation modeling to represent spatiotemporal-unshared sample-specific correlations in the spatiotemporal graph, thus achieving lightweight motion pattern representation and accurate motion prediction.
    \item By mathematically unifying all forms of graph convolutions in human motion prediction, we show that DSTD-GCN improves the representation ability by overcoming the constraints of other graph convolutions.
    \item Extensive experiments show that our proposed DSTD-GCN outperforms other state-of-the-art methods in terms of prediction errors and parameter numbers on three benchmark datasets.
\end{itemize}

\section{Related Work}
  
\subsection{Human Motion Prediction}
Researchers have applied various deep-learning techniques to human motion prediction. CNN-based methods mainly treat a human pose as a pseudo-image. \cite{li2018convolutional,liu2020trajectorycnn}. For example, Liu \textit{et al.} \cite{liu2020trajectorycnn} stacked pose sequences along the channel dimension and extracted multi-level motion features in different CNN layers. RNN-based methods show their capability in modeling temporal motion features and make consistent predictions \cite{fragkiadaki2015recurrent,martinez2017human,li2018convolutional,tang2018long,liu2019towards,li2020dynamic,pavllo2020modeling,liu2021motion,su2021motion,shu2021spatiotemporal,wang2021pvred}. Martinez \textit{et al.} \cite{martinez2017human} proposed a simple yet efficient Seq2Seq model and made predictions based on residual velocities. Wang \textit{et al.} \cite{wang2021pvred} extended this framework by introducing joint velocity and frame position. GAN-based methods make multiple future predictions based on data pattern similarities and produce realistic results \cite{gui2018adversarial,cui2020learning,liu2021aggregated,lyu2021learning}. Lyu \textit{et al.} \cite{lyu2021learning} utilized stochastic differential equations and path integrals.
Transformer-based methods directly model long-range spatial and temporal dependencies \cite{cai2020learning,aksan2021spatio}. Aksan \textit{et al.} \cite{aksan2021spatio} designed spatial and temporal transformers to update spatiotemporal representations simultaneously. Although these methods make remarkable advancements compared with traditional methods, they cannot directly model the natural connectivity between body joints. Natural connectivity is crucial for human motion prediction because human motion follows kinetic chains and trajectory paths, which are constrained by body connections and temporal context. To explicitly model body connections, researchers have turned to GCNs. Graph Convolutions are suitable for processing non-grid and graph-structural data and have been successfully applied to social networks \cite{derr2018signed}, point clouds \cite{qian2021pu}, and traffic prediction \cite{yu2017spatio}. By taking joints as vertices and bones and trajectory connections as edges, a human pose can be naturally represented as a spatiotemporal graph. Recently, many researchers have applied GCs to motion prediction \cite{mao2019learning,cui2020learning,mao2020history,dang2021msr,sofianos2021space,liu2021motion,li2021multiscale}. In this paper, we will only focus on validating our proposed method under conventional offline prediction settings, where the inputs and outputs are all motion sequences. With the success of deep learning in other studies, some works concentrated on extending the prediction settings \cite{dong2022skeleton, li2022online, tevet2023human}. For instance, Dong \emph{et al.} \cite{dong2022skeleton} introduced motion classification as an auxiliary task. Li \emph{et al.} \cite{li2022online} looked into online prediction settings. Tevet \emph{et al.} \cite{tevet2023human} combined natural language information with the Transformer model.
  
\subsection{Graph Convolution Networks in Motion Prediction}
Nearly all GCN-based prediction approaches are developed based on the work of Kipt \text{et al.} \cite{kipf2016semi}, where features are updated in two steps: (1) Feature transformation with a simple linear transformation or a multi-layer perceptron; and (2) Feature aggregation with graph correlations. The graph correlation explicitly depicts connections between body joints, so it is a critical component that distinguishes GCN from other deep learning techniques. Based on the spatiotemporal graph correlation, there are two classifications: (1) \textit{Spatiotemporal-shared/Spatiotemporal-unshared}, which is defined for decomposed spatiotemporal representations and depends on whether the graph correlations are shared by joints in the temporal-unshared spatial graph or the spatial-unshared temporal graph. (2) \textit{Sample-generic/Sample-specific}, which depends on whether graph correlations are adjusted dynamically to different samples.
  
\subsubsection{Spatiotemporal-shared/Spatiotemporal-unshared Methods} For spatiotemporal-shared methods, the spatial/temporal correlations are shared across frames/joints. Most of the previous studies \cite{mao2019learning,mao2020history,cui2020learning,liu2021motion,dang2021msr,li2021multiscale,dong2022skeleton} belong to this type. A limitation of these methods is that they cannot explicitly model changing spatial relations in diverse motion stages and diverse temporal patterns in different joints. Thus, they generally need to stack many graph convolution layers to model complex spatiotemporal features in human motion. For \textit{spatiotemporal-unshared methods}, the spatial/temporal correlations vary from time/joints. Thus, these methods can explicitly depict joint relations in spatiotemporal graphs and precisely model complex spatiotemporal correlations in human motion with far fewer parameters \cite{sofianos2021space}.
  
\subsubsection{Sample-generic/Sample-specific Methods} For sample-generic methods, graph correlations keep unchanged for all samples. Almost all previous studies applied this strategy. Mao \textit{et al.} \cite{mao2019learning} and Dang \textit{et al.} \cite{dang2021msr} directly learned graph correlations from data, while Li \textit{et al.} \cite{li2020dynamic} initialized the trainable correlation as a predefined graph. Cui \textit{et al.} \cite{cui2020learning} and Liu \textit{et al.} \cite{liu2021motion} combined the previous two correlations and proposed a semi-constrained graph correlation. For \textit{sample-specific methods}, graph correlations can be adjusted to each sample. To our best knowledge, there is no application of sample-specific graph convolutions to motion prediction. We are the first to model \emph{sample-specific spatiotemporal-unshared graph convolutions} in motion prediction.

\section{Methodology}

The proposed dynamic SpatioTemporal Decompose Graph Convolution (DSTD-GC) is introduced in this section. First, human motion prediction and some graph-related definitions are presented. Then, all graph convolutions in motion prediction are summarized in a unified form. Next, constrained dynamic correlation modeling and Dynamic Spatiotemporal Graph Convolution are described. Finally, the detailed model architecture is introduced.
  
\subsection{Preliminaries}
  
\subsubsection{Problem Definition} Human motion prediction is to predict $L$ future human pose frames with $K$ historical observations. The entire sequence length is $T = K + L$. As a human pose is represented by $J$ joints with $D-$dimensional spatial information, the human pose at time $t$ is denoted as $X_t \in \mathbb{R}^{J \times D}$. The historical observations are formulated as $X_{1:K} = \{X_1,\cdots,X_K\}$. Our goal is to predict the future human motion $\tilde{X}_{K+1:K+L} = \{\tilde{X}_{K+1}, \cdots,\tilde{X}_{K+L}\}$, where the corresponding ground truth is denoted as $X_{K+1:K+L} = \{X_{K+1},\cdots,X_{K+L}\}$.
  
\subsubsection{Notations}\label{sec:notations} The spatiotemporal graph and its decomposition are formally defined as follows.
  
\emph{Spatiotemporal Graph Representation.} As shown in Fig.~\ref{fig:motivation} (a), a human motion sequence with $T$ frames is presented as a spatiotemporal graph $\mathcal{G}^{st}=(\mathcal{V}^{st}, \mathcal{E}^{st})$, where $\mathcal{V}^{st} \in \mathbb{R}^{JT}$ is the set of all joints across frames, and $\mathcal{E}^{st}$ is the set of spatiotemporal edges. $\boldsymbol{A}^{st} \in \mathbb{R}^{JT \times JT} $ is the spatiotemporal adjacency matrix, which represents the correlations between the vertices in the graph. However, it takes many parameters to directly store and learn spatiotemporal correspondences with space complexity of $O((JT)^2)$. Therefore, we further decompose a spatiotemporal graph into a unique pair of $T$ spatial graphs across frames and $J$ temporal graphs across joints. The spatial graphs are denoted as $\mathcal{G}^{s} = (\mathcal{V}^{s}, \mathcal{E}^{s})$ where $\mathcal{V}^{s} \in \mathbb{R}^{T \times J}$ is the set of joint vertices. $\mathcal{E}^{s}$ is the set of spatial edges, and it is formulated as a temporal-unshared spatial adjacency matrix\footnote{For clarity, the third-order tensor is denoted as a matrix, and the second-order tensor is denoted as a vanilla matrix.} $\boldsymbol{A}^{s} \in \mathbb{R}^{T \times J \times J}$. When spatial correlations are shared across frames, the spatial adjacency matrix is degraded into a vanilla form and is represented by a matrix from $\mathbb{R}^{J \times J}$. Similarly, the temporal graphs are denoted as $\mathcal{G}^{t} = (\mathcal{V}^{t}, \mathcal{E}^{t})$ where $\mathcal{V}^{t} \in \mathbb{R}^{J \times T}$ is the set of trajectory vertices. $\mathcal{E}^{t}$ is the set of temporal edges, and it is formulated as a spatial-unshared temporal adjacency matrix $\boldsymbol{A}^{t} \in \mathbb{R}^{J \times T \times T}$.
  
\emph{Spatiotemporal-equivalence.} Spatiotemporal-equivalence is defined for two functions on spatial graphs and corresponding temporal graphs. Formally, function $\mathcal{F}^{s}$ on $\mathcal{G}^{s}$ and function $\mathcal{F}^{t}$ on the corresponding $\mathcal{G}^{t}$ are spatiotemporal-equivalent if and only if $\mathcal{F}^{s}$ is equivalent to $\mathcal{F}^{t}$ after switching all operations for the time and space dimensions, and vice versa. The graph convolutions with spatiotemporal-equivalent operations are called spatiotemporal-equivalent graph convolutions. In the following, we only introduce the function on spatial graphs for a spatiotemporal-equivalence pair.
  
\subsubsection{Spatiotemporal Decompose Graph Convolutions}\label{sec:stdgc_intro} We introduce different graph convolutions by taking the example of the feature updating process for joint $q$ at frame $n$. The spatial and temporal neighbor indices of the selected joint are denoted as $p$ and $m$, respectively. For a typical \hypertarget{stgc_intro}{spatiotemporal graph convolution (ST-GC)}, the output feature $\mathbf{y}_{qn}$ is obtained by:
\begin{equation}
    \mathbf{y}_{qn} = \sum_{p}^{J} \sum_{m}^{T} a^{st}_{(pm)(qn)} \mathbf{x}_{pm} \boldsymbol{W},
    \label{eq:st-gc}
\end{equation}
where $\mathbf{x}_{pm} \in \mathbb{R}^{C}$ is the input feature from joint $p$ of frame $m$, $\mathbf{y}_{qn} \in \mathbb{R}^{C'}$ is the output feature, $a^{st}_{(pm)(qn)}$ is a correlation strength in the spatiotemporal adjacency matrix $\boldsymbol{A}^{st} \in \mathbb{R}^{JT \times JT}$, and $\boldsymbol{W} \in \mathbb{R}^{C \times C'}$ is a trainable parameter for feature transformation\footnote{The value of $\boldsymbol{W}$ varies in different graph convolutions.}.
  
With the graph decomposition introduced above, two graph GCs are defined: spatial graph convolution (S-GC) and temporal graph convolution (T-GC):
\begin{equation}
    \mathbf{x}^{s}_{qn}= \sum_{p}^{J} a^{s}_{npq} \mathbf{x}_{pn} \boldsymbol{W}_{1},
    \label{eq:sgcn}
\end{equation}
\begin{equation}
    \mathbf{x}^{t}_{qn} = \sum_{m}^{T} a^{t}_{qmn} \mathbf{x}_{qm} \boldsymbol{W}_{2},
    \label{eq:tgcn}
\end{equation}
where $\boldsymbol{W}_{1}$ and $\boldsymbol{W}_{2}$ are feature transformation matrices from $\mathbb{R}^{C \times C^{'}}$ and $\mathbb{R}^{C^{'} \times C^{'}}$.

Specifically, the S-GC models frame-wise spatial correlations, while T-GC models joint-wise temporal correlations. Then, we propose \textit{\hypertarget{stdgc_intro}{SpatioTemporal Decompose Graph Convolution (STD-GC)}} by alternatively stacking a S-GC (Eq. \ref{eq:sgcn}) and a T-GC (Eq. \ref{eq:tgcn}):
\begin{equation}
    \mathbf{y}_{qn} = \sum_{m}^{T} a^{t}_{qmn} (\sum_{p}^{J} a^{s}_{npq} \mathbf{x}_{pm} \boldsymbol{W}_{1}) \boldsymbol{W}_{2},
    \label{eq:stdgcn}
\end{equation}
where $a^{s}_{npq}$ and $a^{t}_{qmn}$ are from a spatial adjacency matrix $\boldsymbol{A}^{s}$ and a temporal adjacency matrix $\boldsymbol{A}^{t}$. STD-GC decomposes ST-GC on a spatiotemporal graph into S-GC on spatial graphs and T-GC on temporal graphs. Then, spatiotemporal feature modeling is performed by stacking these two GCs. The capability of spatiotemporal modeling is not affected by the stacking order of S-GC and T-GC. Meanwhile, STD-GC is equivalent to STS-GC \cite{sofianos2021space}. All of these will be demonstrated with experiments in Sect. \ref{sec:gc_compare}. Furthermore, STD-GC becomes \hypertarget{vstdgc_intro}{vanilla STD-GC (VSTD-GC)} when the spatial and temporal adjacency matrices are in vanilla form.
  
In sample-generic methods, $a^{s}_{npq}$ and $a^{t}_{qmn}$ are set based on prior knowledge or defined as trainable parameters. In sample-specific methods, $a^{s}_{npq}$ and $a^{t}_{qmn}$ are generated by a model according to the input features.

\subsection{Analysis of Graph Convolutions on Spatiotemporal Graphs}
  
By reformulating several graph convolutions on spatiotemporal graphs into a uniform form, we evaluate their representation capabilities in human motion prediction. First, the spatiotemporal graph decomposition constraint is introduced, and then several graph convolutions are evaluated based on their spatial and temporal modeling capability. We focus exclusively on the correlation between distinct graph convolutions because the correlation dominates feature aggregation and is critical for obtaining representative spatiotemporal features. Based on the formulation in Sect. \ref{sec:stdgc_intro}, we adds $(i)$ in the superscript to indicate different samples.

By comparing ST-GC (Eq. \ref{eq:st-gc}) and STD-GC (Eq. \ref{eq:stdgcn}), it can be seen that the spatiotemporal correlations can be represented by a combination of spatial and temporal correlations with the following constraints: \\
\textit{\hypertarget{constraint1}{Constraint 1}: $a^{st(i)}_{(pm)(qn)}$ equals to the product of $a^{s(i)}_{npq}$ and $a^{t(i)}_{qmn}$}. \\
where $a^{st(i)}_{(pm)(qn)}$ is the spatiotemporal correlation strength, $a^{s(i)}_{npq}$ is the corresponding spatial correlation strength, and $a^{t(i)}_{qmn}$ is the corresponding temporal correlation strength. 
  
With the decomposition constraint, we further reformulate decomposed ST-GC from the spatial and temporal graph perspectives. These two views are presented as follows:
\begin{equation}
    \mathbf{y}^{(i)}_{qn} = \sum_{p}^{J} a^{s(i)}_{npq} \sum_{m}^{T} \mathbf{x}^{(i)}_{pm} (a^{t(i)}_{qmn} \boldsymbol{W})
    \label{eq:gc-s}
\end{equation}
\begin{equation}
    \mathbf{y}^{(i)}_{qn} = \sum_{m}^{T} a^{t(i)}_{qmn} \sum_{p}^{J} \mathbf{x}^{(i)}_{pm} (a^{s(i)}_{npq} \boldsymbol{W})
    \label{eq:gc-t}
\end{equation}
where $a^{s(i)}_{npq}$ and $a^{t(i)}_{qmn}$ are the spatial and temporal correlation weights for the input sample. Meanwhile, we combine correlation strengths and feature transformation weights into a generalized weight matrix $a^{s(i)}_{npq} \boldsymbol{W}(a^{s(i)}_{npq} \boldsymbol{W})$. Thus, the feature updating process consists of feature transformation with the generalized weight matrix and feature aggregation with frame-wise spatial or joint-wise temporal correlations. With these two formulations, we analyze different graph GCs based on the spatial and temporal correlations for feature aggregation.

\subsubsection{Sample-generic Spatiotemporal-shared GCs} Most recent studies \cite{yan2018spatial,mao2019learning,mao2020history,cui2020learning,dang2021msr,li2021multiscale} adopted spatiotemporal-shared GCs, where joint vertices from all frames share one spatial correlation and trajectory vertices for all joints share one temporal correlation.
The whole GC process can be formulated as:
\begin{equation}
\mathbf{y}^{(i)}_{qn} = \sum_{p}^{J} \sum_{m}^{T} a^{s}_{pq} a^{t}_{mn} \mathbf{x}^{(i)}_{pm} \boldsymbol{W},
\label{eq:s/tgc}
\end{equation}
\begin{equation}
\mathbf{y}^{(i)}_{qn} = \sum_{p}^{J} a^{s}_{pq} \sum_{m}^{T} \mathbf{x}^{(i)}_{pm} (a^{t}_{mn} \boldsymbol{W}),
\label{eq:s/tgc-s}
\end{equation}
\begin{equation}
\mathbf{y}^{(i)}_{qn} = \sum_{m}^{T} a^{t}_{mn} \sum_{p}^{J} \mathbf{x}^{(i)}_{pm} (a^{s}_{pq} \boldsymbol{W}),
\label{eq:s/tgc-t}
\end{equation}
where $a^{s}_{pq}$ is the shared spatial correlation strength and $a^{t}_{mn}$ is the shared temporal correlation strength. They are from the vanilla form of the spatial adjacency matrix $\boldsymbol{A}^{s}$ and the temporal adjacency matrix $\boldsymbol{A}^{t}$, respectively. Eq. \ref{eq:s/tgc-s} and \ref{eq:s/tgc-t} are the reformulations of Eq. \ref{eq:s/tgc} in the spatial graph's view and temporal graph's view. Besides, these two correlation strengths keep unchanged across input samples. Therefore, the sample-generic spatiotemporal-shared GCs are subject to the following constraints: \\
\textit{\hypertarget{constraint2}{Constraint 2}: $a^{s(i)}_{n_{1}pq}$ and $a^{s(i)}_{n_{2}pq}$} are forced to be the same, $a^{t(i)}_{q_{1}mn}$ and $a^{t(i)}_{q_{2}mn}$ are forced to be the same. \\
\textit{\hypertarget{constraint3}{Constraint 3}: $a^{s(i_1)}_{npq}$ and $a^{s(i_2)}_{npq}$} are forced to be the same, $a^{t(i_1)}_{qmn}$ and $a^{t(i_2)}_{qmn}$ are forced to be the same. \\
Note that $i_{1}$ and $i_{2}$ are indices of different input samples; $n_{1}$ and $n_{2}$ are indices of different joints; $q_{1}$ and $q_{2}$ are indices of different frames.

\subsubsection{Sample-generic Spatiotemporal-unshared GCs} The only difference between sample-generic spatiotemporal-shared GCs and sample-generic spatiotemporal-unshared GCs is that the spatiotemporal-unshared ones adopt unshared spatial and temporal correlations. The sample-generic spatiotemporal-unshared GCs can be formulated as:
\begin{equation}
\mathbf{y}^{(i)}_{qn} = \sum_{p}^{J} \sum_{m}^{T} a^{s}_{npq} a^{t}_{qmn} \mathbf{x}^{(i)}_{pm} \boldsymbol{W},
\label{eq:stgc-static}
\end{equation}
\begin{equation}
\mathbf{y}^{(i)}_{qn} = \sum_{p}^{J} a^{s}_{npq} \sum_{m}^{T} \mathbf{x}^{(i)}_{pm} (a^{t}_{qmn} \boldsymbol{W}),
\label{eq:stgc-s-static}
\end{equation}
\begin{equation}
\mathbf{y}^{(i)}_{qn} = \sum_{m}^{T} a^{t}_{qmn} \sum_{p}^{J} \mathbf{x}^{(i)}_{pm} (a^{s}_{npq} \boldsymbol{W}),
\label{eq:stgc-t-static}
\end{equation}
where $a^{s}_{npq}$ is frame-wise spatial correlation strength and $a^{t}_{qmn}$ is joint-wise temporal correlation strength. They are from the spatial adjacency matrix and the temporal adjacency matrix, respectively. With the two unshared correlations, sample-generic spatiotemporal-unshared GCs capture evolving spatial relations between joints at different action stages and diverse temporal relations for different body parts. Thus, they generally outperform spatiotemporal-shared GCs. However, these correlations are optimized across all data and may not be optimized for individual action sequences. From the formula's view, sample-generic spatiotemporal-unshared GCs still suffer from \textit{Constraint \hyperlink{constraint3}{3}} but relax \textit{Constraint \hyperlink{constraint2}{2}} into the following constraint: \\
\textit{Constraint \hypertarget{constraint4}{4}: $a^{s(i)}_{n_{1}pq}$ and $a^{s(i)}_{n_{2}pq}$ differ by a scaling factor, $a^{t(i)}_{q_{1}mn}$ and $a^{t(i)}_{q_{2}mn}$ differ by a scaling factor.} 

\subsubsection{Sample-specific Spatiotemporal-unshared GCs} In comparison to sample-generic spatiotemporal-unshared GCs, the sample-specific ones infer dynamic correlations between vertices and thus have a higher representation capacity. The formulation for these GCs is as follows:
\begin{equation}
\mathbf{y}^{(i)}_{qn} = \sum_{p}^{J} \sum_{m}^{T} a^{s(i)}_{npq} a^{t(i)}_{qmn} \mathbf{x}^{(i)}_{pm} \boldsymbol{W},
\label{eq:stgc-dynamic}
\end{equation}
\begin{equation}
\mathbf{y}^{(i)}_{qn} = \sum_{p}^{J} a^{s(i)}_{npq} \sum_{m}^{T} \mathbf{x}^{(i)}_{pm} (a^{t(i)}_{qmn} \boldsymbol{W}),
\label{eq:stgc-s-dynamic}
\end{equation}
\begin{equation}
\mathbf{y}^{(i)}_{qn} = \sum_{m}^{T} a^{t(i)}_{qmn} \sum_{p}^{J} \mathbf{x}^{(i)}_{pm} (a^{s(i)}_{npq} \boldsymbol{W}),
\label{eq:stgc-t-dynamic}
\end{equation}
where both $a^{s(i)}_{npq}$ and $a^{t(i)}_{qmn}$ are adjustable to the $i$-th input sample. Based on these formulations, sample-specific spatiotemporal-unshared GCs relax both \textit{Constraint \hyperlink{constraint2}{2}} and \hyperlink{constraint3}{\textit{3}}. Specifically, they relax \textit{Constraint \hyperlink{constraint3}{2}} into \textit{Constraint \hyperlink{constraint4}{4}} and relax \textit{Constraint \hyperlink{constraint3}{3}} into the following constraint:\\
\textit{\hypertarget{constraint5}{Constraint 5}: $a^{s(i_1)}_{npq}$ and $a^{s(i_2)}_{npq}$ differ by a scaling factor, $a^{t(i_1)}_{qmn}$ and $a^{t(i_2)}_{qmn}$ differ by a scaling factor.}

From the above analysis, it can be seen that sample-specific spatiotemporal-unshared GCs are the least constrained. They are also the theoretically strongest decomposed GCs for spatiotemporal feature representation. This will be demonstrated in Sect. \ref{sec:gc_compare} by comparing different graph convolutions.

\begin{figure}[htbp]
  \centering
  \includegraphics[width=0.48\textwidth]{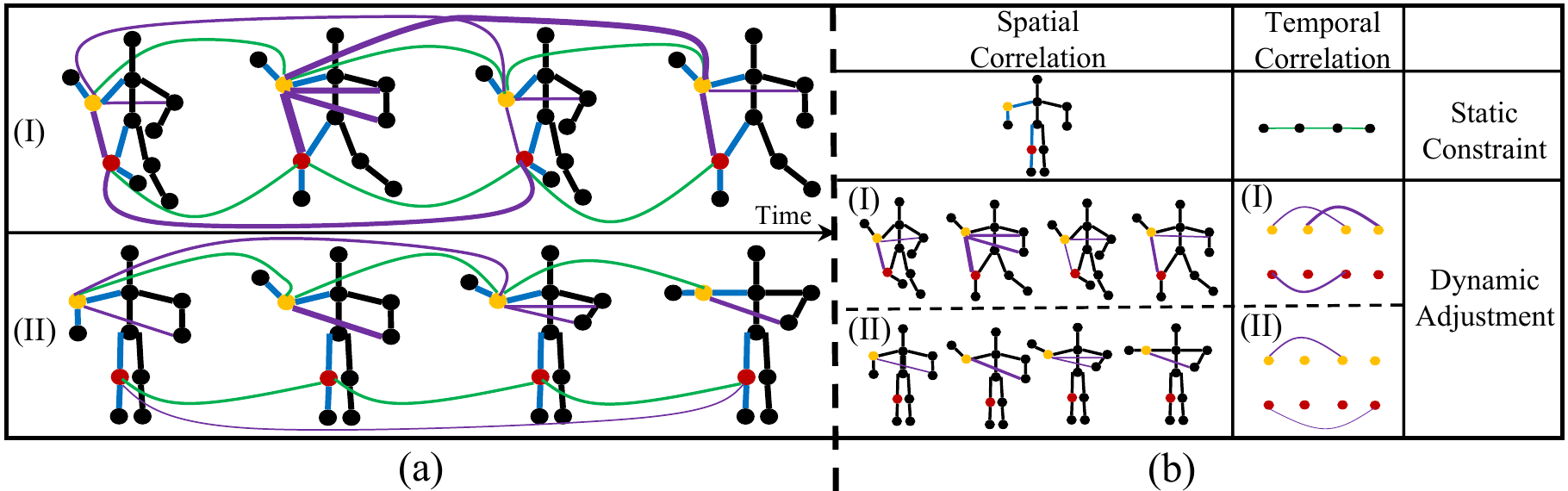}
  \vspace{-0.2cm}
  \caption{Constrained dynamic correlations. (a) Spatial and temporal correlations in (I) walking and (II) directing traffic. (b) The spatial/temporal correlations in (a) can be divided into static constraints (blue/green) and dynamic adjustments (purple). The constraints are shared by all frames/joints and stay the same for all samples, while the adjustments are unique to different frames/joints and vary according to each sample. We use the yellow elbow and the red knee joints as an example. The line indicates spatial correlations, while the curve represents temporal correlations}
  \label{fig:constrained_dynamic_correlation}
  \vspace{-0.5cm}
\end{figure}

\begin{figure*}[htbp]
  \centering
  \includegraphics[width=1\textwidth]{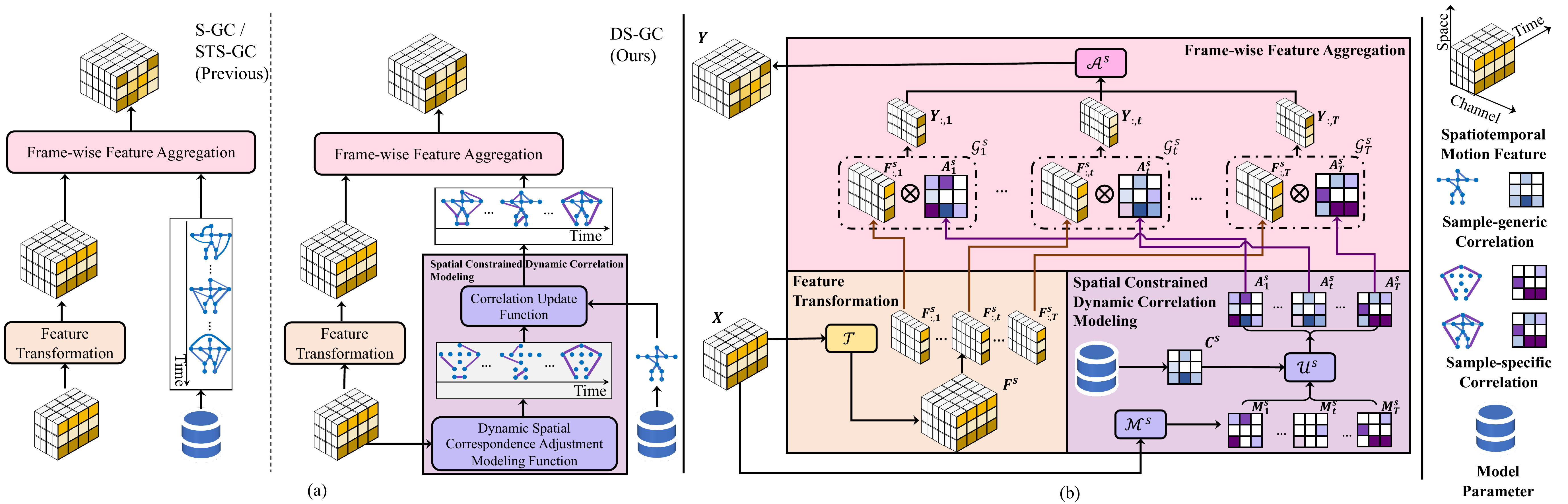}
  \vspace{-0.4cm}
  \caption{(a) The overview and comparison of spatiotemporal-unshared graph convolution by taking spatial modeling as an example. Previous methods store the frame-wise spatial correspondence as a model parameter, while our approach only saves information shared by all frames and extracts frame-wise adjustments from the input sequence. (b) The framework of dynamic spatial graph convolution. Feature transformation projects input spatiotemporal features into high-level representations. Spatial constrained dynamic correlation modeling refines spatial-constrained adjacency matrix with sample-specific frame-wise adjustments. The adjusted adjacency matrix guides feature aggregation for the high-level representation.} 
  \label{fig:dsgc}
  \vspace{-0.2cm}
\end{figure*}

\subsection{Dynamic SpatioTemporal Decompose Graph Convolutions}
Below, our proposed \textit{\hypertarget{dstdgc_intro}{Dynamic SpatioTemporal Decompose Graph Convolution (DSTD-GC)}} is introduced, which extends \hyperlink{stdgc_intro}{STD-GC} with Dynamic Spatial Graph Convolutions (DS-GC) and its spatiotemporal-equivalent counterpart Dynamic Temporal Graph Convolutions (DT-GC). The main difference between DSTD-GC and STD-GC is that the former GC equips constrained dynamic correlation modeling, which is the key component for powerful and lightweight representations. In this section, constrained dynamic correlation modeling is introduced, and then the details of DS-GC are presented.

\subsubsection{Constrained Dynamic Correlation Modeling}
The motivation behind constrained dynamic correlation modeling is that the correlations between joints are both constrained and dynamic. Specifically, the correlations between joints are constrained by the inherent human body structure, consistent trajectory connections, and learned motion prior. Meanwhile, these correlations are dynamically changed based on different motion sequences. To show these two characteristics, we utilize the correlations of the elbow and knee joints in the walking and directing traffic scenarios as an example (Fig. \ref{fig:constrained_dynamic_correlation}). The elbow and knee movements generally follow the motion chain of the human body and common trajectory patterns, so their spatial/temporal correlation presents static constraints, which are shared by all frames/joints and stay fixed under these two scenarios. Meanwhile, compared with the dynamic spatial and temporal adjustments in the directing traffic scenario, the spatial adjustment strengths between the elbow joint and the knee joint and temporal adjustment strengths in the knee joint are generally stronger in the walking scenario. These dynamic adjustments refine static spatial/temporal constraints at each frame/joint and vary from different inputs.
 
Inspired by this, we present constrained dynamic correlation modeling to directly represent the aforementioned static constraints and dynamic adjustments. Taking the spatial modeling from the purple box in Fig. \ref{fig:dsgc} (a) as an example, the static constraints are parameterized as a shared adjacency matrix, while the dynamic adjustments are extracted by the dynamic spatial correlation adjustment modeling function. This function takes the spatiotemporal motion feature as inputs and outputs frame-wise spatial correspondence adjustments that update static constraints for frame-wise feature aggregation.

\subsubsection{Dynamic Spatial Graph Convolutions}
The general framework of DS-GC is illustrated in Fig. \ref{fig:dsgc}(b). Specifically, DS-GC consists of three parts: (1) Feature Transformation; (2) Spatial constrained dynamic correlation modeling; (3) Frame-wise feature aggregation. DS-GC takes spatiotemporal motion feature $\boldsymbol{X} \in \mathbb{R}^{J \times T \times C}$ and constrained spatial adjacency matrix $\boldsymbol{A}^{s} \in \mathbb{R}^{J \times J}$ as inputs, and then it outputs $\boldsymbol{Y} \in \mathbb{R}^{J \times T \times C^{'}}$. This section will only introduce the functions in DS-GC, and the implementations will be presented in Sect. \ref{sec:model_arch}. 

\emph{Feature Transformation:} As shown in the yellow block of Fig. \ref{fig:dsgc} (b), feature transformation is accomplished by transformation function $\mathcal{T}$. Here, we set the function as a linear transformation function for clarity, while other functions, such as a multi-layer perception, can also be used. The function is defined as:
\begin{equation}
    \boldsymbol{F}^{s} = \mathcal{T}(\boldsymbol{X}) \triangleq \boldsymbol{X} \boldsymbol{W},
\end{equation}
where $\boldsymbol{F}^{s} \in \mathbb{R}^{J \times T \times C^{'}}$ is the high-level representation.

\emph{Spatial Constrained Dynamic Correlation Modeling:} The spatial constrained dynamic correlation modeling is shown in the purple part of Fig. \ref{fig:dsgc} (b). Here, the static constraints are parameterized as a vanilla adjacency matrix $\boldsymbol{C}^{s} \in \mathbb{R}^{1 \times J \times J}$, which is set as a model parameter and optimized across all data. Meanwhile, the dynamic correspondence adjustments $\boldsymbol{M}^{s} \in \mathbb{R}^{T \times J \times J}$ are inferred from the input sample to capture frame-wise spatial correspondence adjustments. Specifically, $\boldsymbol{M}^{s}$ is extracted by the dynamic spatial correspondence adjustment modeling function $\mathcal{M}^{s}$, which traverses all pairs of joints and extracts mutual relations. The formulation of this function for a joint pair $(p, q)$ is presented below:
\begin{equation}
    \boldsymbol{M}^{s}= \mathcal{M}^{s}(\mathbf{x}_{p}, \mathbf{x}_{q}) \triangleq \operatorname{MLP}(\theta(\boldsymbol{x}_{p}) ||^{s}_{p} \phi(\mathbf{x}_{q})),
    \label{eq:dsgc-sdm}
\end{equation}
where $(\mathbf{x}_{p},\mathbf{x}_{q})$ are from the input feature $\boldsymbol{X}$ of the joint pair $(p, q)$, and $||^{s}_{p}$ is the pair-wise concatenation along the joint dimension. $\theta$ and $\phi$ are two linear transformation functions that project the graph vertex features into low-dimensional representations to reduce computational cost. $\operatorname{MLP}$ is a multi-layer perceptron, we utilize $\operatorname{MLP}$ because it can model complex spatial correlations. Note that $\boldsymbol{M}^{s}$ is not required to be symmetric, which benefits correlation modeling by increasing analysis flexibility and enhancing representative ability. Finally, the output spatial constrained dynamic correlation, $\boldsymbol{A}^{s}$, is obtained by a correlation updating function $\mathcal{U}^{s}_{t}$. This function adjusts the static constraints $\boldsymbol{C}^{s}$ with dynamic adjustments $\boldsymbol{M}^{s}$:
\begin{equation}
    \boldsymbol{A}^{s} = \mathcal{U}^{s}_{t}(\boldsymbol{C}^{s}, \boldsymbol{M}^{s}) \triangleq \boldsymbol{C}^{s} + \alpha \cdot \boldsymbol{M}^{s},
    \label{eq:dsgc-correlation}
\end{equation}
where $\alpha$ is a learnable parameter for controlling the adjustment intensity.

\emph{Frame-wise Feature Aggregation:} As shown in the pink block of Fig. \ref{fig:dsgc} (b), given the spatial constrained dynamic correlations $\boldsymbol{A}^{s}$ and high-level features $\boldsymbol{F}^{s}$, the output is obtained with the frame-wise aggregation function $\mathcal{A}^{s}_{t}$. Specifically, $\mathcal{A}^{s}_{t}$ constructs a spatial graph along the temporal dimension for the updated correlation $\boldsymbol{A}^{s}_{t}$ and feature $\boldsymbol{F}^{s}_{:, t}$, where $\boldsymbol{A}^{s}_{t}$ and $\boldsymbol{F}^{s}_{:, t}$ are from the $t$-th frame of $\boldsymbol{A}^{s}$ and $\boldsymbol{F}^{s}$, respectively. The output feature $\boldsymbol{Y} \in \mathbb{R}^{J \times T \times C^{'}}$ is obtained by:
\begin{equation}
    \boldsymbol{Y} = \mathcal{A}^{s}(\boldsymbol{F}^{s}, \boldsymbol{M}^{s}) \triangleq [\boldsymbol{F}^{s}_{:, 1} \boldsymbol{M}^{s}_{1} ||_{t} \cdots ||_{t} \boldsymbol{F}^{s}_{:, T} \boldsymbol{M}^{s}_{T}],
\end{equation}
where $||_{t}$ is a concatenation function along the frame dimension.

Based on the above three steps, DS-GC is formulated as:
\begin{equation}
    \boldsymbol{Y} = \mathcal{A}^{s}(\mathcal{T}(\boldsymbol{X}), \mathcal{U}^{s}(\boldsymbol{C}^{s}, \mathcal{M}^{s}(\boldsymbol{X}))).
\end{equation}

Since DT-GC is spatiotemporal-equivalent to DS-GC, DT-GC can be directly obtained by replacing all operations in DS-GC with their spatiotemporal-equivalent counterparts:
\begin{equation}
    \boldsymbol{Y} = \mathcal{A}^{t}(\mathcal{T}(\boldsymbol{X}), \mathcal{U}^{t}(\boldsymbol{C}^{t}, \mathcal{M}^{t}(\boldsymbol{X}))).
\end{equation}

Note that in DS-GC, the spatial constrained dynamic correlations are sample-specific and unshared in Eq. \ref{eq:dsgc-sdm}. Similarly, the temporal constrained dynamic correlations from DT-GC are sample-specific and unshared. Therefore, the DSTD-GC, which combines DS-GC and DT-GC for spatiotemporal modeling, belongs to the sample-specific spatiotemporal-unshared GC. The detailed derivation is presented in Appendix \hyperlink{sec:appendix-a}{A}.
  
\subsubsection{Discussions}\label{sec:dstdgc-dis} Fig. \ref{fig:dsgc} (a) shows the comparison between STS-GC/S-GC and DS-GC in spatial modeling. It is found that DSTD-GC has three advantages: (1) \textit{Parameter-saving spatiotemporal-unshared representation}: To accomplish unshared spatial correlation modeling, STS-GC/S-GC directly parameterizes the correspondence as a spatial adjacency matrix $\boldsymbol{A}^{s}$ in the GC, where the space complexity is $O(JT^{2})$. Conversely, DS-GC accomplishes unshared spatial modeling with space complexity of $O(J^{2})$ by learning a vanilla adjacency matrix and a function with negligible parameters. Similarly, the space complexity of spatial-unshared temporal correlation modeling is $O(J^{2}T)$ for STS-GC/T-GC and $O(T^{2})$ for DT-GC. As the joint number is always proportional to the sequence length, the overall space complexity for STS-GC and STD-GC is $O(T^{3})$ and that for DSTD-GC is only $O(T^{2})$. (2) \textit{Joint optimization}: DSTD-GC learns common motion patterns in a vanilla matrix and a function to extract frame-wise/joint-wise correlation variations. With fewer parameters to learn, this mechanism has less optimization difficulty than directly learning optimal spatial/temporal correlations through an adjacency matrix. (3) \textit{Sample-specific modeling}. DSTD-GC models dynamic spatiotemporal correspondence adjustments in different motion inputs. These adjustments reflect sample-specific motion patterns and contribute to accurate motion predictions. In addition to these advantages, our DSTD-GC introduces little computation overhead. The analysis of the time complexity of DSTD-GC and STS-GC is presented in Appendix \hyperlink{sec:appendix-b}{B}. Both graph convolutions have the same time complexity of $O(CT^2(C+T))$.

\subsection{Model Architecture} \label{sec:model_arch}

Based on DSTD-GC, we propose a lightweight and powerful model called DSTD-GCN, which is built upon the prediction framework proposed by \cite{mao2019learning}. In this section, the prediction framework is introduced first, and then the implementation details of DSTD-GCN are given.

\begin{figure}[!ht]
  \centering
  \includegraphics[width=0.48\textwidth]{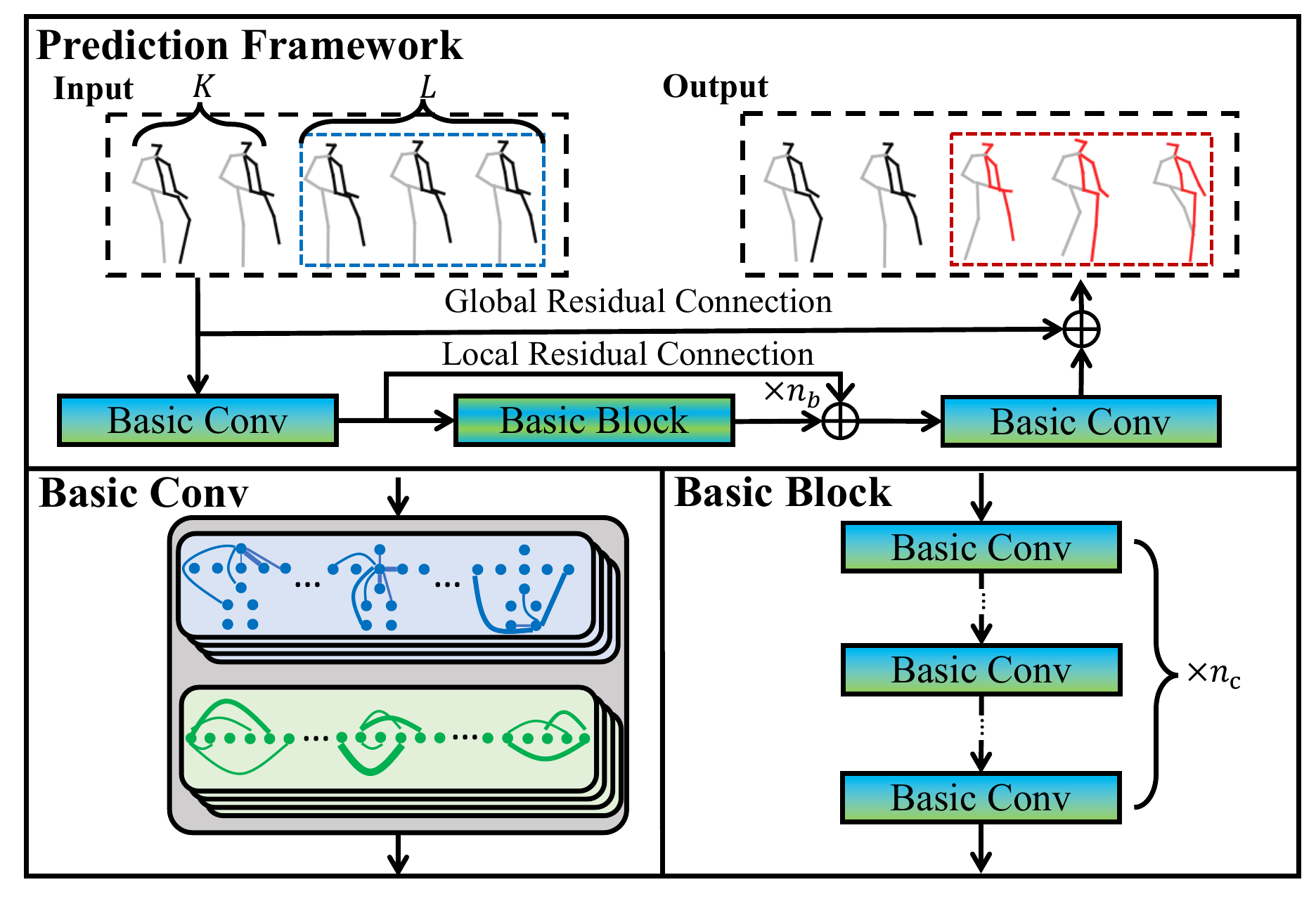} 
  \vspace{-0.2cm}
  \caption{The problem reformulation and prediction framework. The human poses in the blue dashed box are duplicate inputs, while the poses in the red box are target predictions. The prediction framework contains two Basic Conv units and $n_b$ basic blocks. The two Basic Conv units perform transformation between raw motion sequences and high-dimensional spatiotemporal features. Each basic block contains $n_c$ Basic Conv units for feature updating.} 
  \label{fig:framework}
  \vspace{-0.4cm}
\end{figure}

\subsubsection{Prediction Framework} We duplicate the last input human pose for $L$ times and formulate a new input motion sequence $X_{1:K+L}$. Then the original prediction task is reformulated into predicting the residual motion sequence between $\tilde{X}_{1:K+L}$ and the corresponding ground truth $X_{1:K+L}$. Based on the problem reformulation, the prediction framework is proposed as shown in Fig. \ref{fig:framework}. The framework contains a Basic Conv unit to encode motion input to high-level representations, $n_b$ Basic Blocks for spatiotemporal representation modeling, and a Basic Conv unit to decode predictions from motion features. The Basic Conv unit is a single convolution layer like CNN and ST-GC. It stacks $n_c$ layers to constitute a basic block. Besides, the global residual connection enforces the residual learning, while the local residual connection stabilizes the training process. Recent studies \cite{mao2019learning,cui2020learning,dang2021msr} prove the effectiveness of this framework. With this framework, the input and output human poses can be modeled as a whole sequence, and multi-level associated spatiotemporal correlations can be modeled, while the previous frameworks \cite{li2020dynamic,sofianos2021space} restrict high-level feature interactions between the input and output feature. This framework will be used to build our model and compare different GCs.

\begin{figure}[!ht]
  \centering
  \vspace{-0.2cm}
  \includegraphics[width=0.48\textwidth]{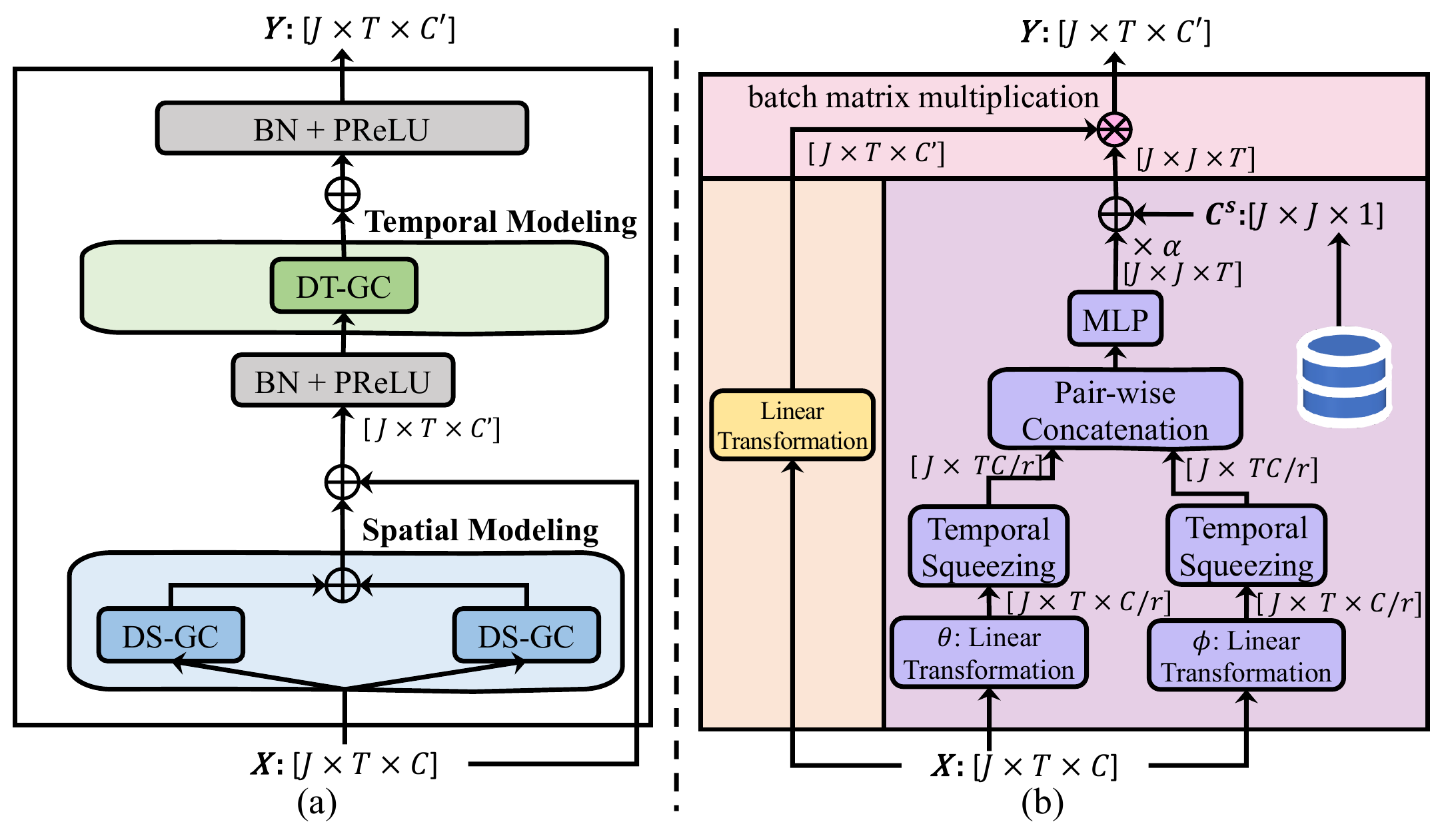}
  \vspace{-0.3cm}
  \caption{(a) The Basic Conv unit of DSTD-GCN. Spatial modeling aims to adjust the natural and semantic connection constraints separately with two DS-GCs, while temporal modeling refines the context adjacency constraint with a DT-GC. (b) The implementation of DS-GC. The operations with different colors indicate different functions.}
  \label{fig:details}
\end{figure}

\subsubsection{Model Implementation} The entire network adopts the prediction framework and stacks five basic blocks. Each basic block contains one Basic Conv unit. The Basic Conv unit of DSTD-GCN is shown in Fig. \ref{fig:details} (a), which contains two parallel DS-GCs for spatial modeling and a DT-GC for temporal modeling. Specifically, for DS-GC, the modeling function $\mathcal{M}^{s}$ takes $X \in \mathbb{R}^{J \times T \times C}$ as input. Two linear transformation functions $\theta$ and $\phi$ with a reduction rate of $r$ are used to transform $\boldsymbol{X}$ into a neatly compact representation and then squeeze the temporal dimension into the channel dimension. After this, a pair-wise concatenation and a multi-layer perceptron are adopted to extract spatial adjustments for each frame based on Eq. \ref{eq:dsgc-sdm}. These dynamic spatial adjustments update the static constraints $\mathbf{C}^{s}$ to obtain the sample-specific frame-wise spatial correlation. Finally, batch matrix multiplication is used for frame-wise feature aggregation to obtain the output $\mathbf{Y}$.

We further consider prior knowledge for the static constrained correlation. As Cui \textit{et al.} \cite{cui2020learning} pointed out that a restricted prior graph correlation will reduce flexibility, we consider a simple solution here. We adopt a trainable constrained adjacency matrix and initialize it according to prior knowledge, where the weight is set to 1 for correlated joint pairs and 0 for other pairs. For spatial correspondence, we consider two types of joint relations. It is assumed that these two relations have independent influence and should be modeled separately. The first relation is the body connection, where a joint is correlated with its physically connected neighbors. The second relation is the semantic correlation. Since the arms and legs have a large variability and are hard to predict, we connect vertices in the same arm/leg. Meanwhile, body symmetry is considered, and the arm or leg is connected to its mirror counterpart. For temporal constraints, we only consider temporal context and connect every frame to its adjacent one.

\section{Experiments}

To evaluate the effectiveness of our proposed model, experiments are conducted on three standard benchmark motion capture datasets, including Human3.6M \cite{ionescu2013human3}, CMU Mocap, and 3DPW \cite{marcard2018recovering} datasets. Meanwhile, experiments are designed to answer the following questions: \textit{(1) what is the performance of DSTD-GC as compared to other graph convolutions? (2) what is the performance of our proposed DSTD-GCN as compared to other state-of-art motion prediction methods? (3) what is the influence of different components and what findings can be obtained from the experimental results?}

\subsection{Dataset Settings}

\subsubsection{Human3.6M} Human3.6M \cite{ionescu2013human3} is a widespread benchmark dataset for evaluating motion prediction. It contains 15 actions performed by 7 actors. Following the data preprocessing procedure of \cite{martinez2017human,mao2019learning}, the original data in the exponential mapping format is transformed into 3D joint coordinate space, and a single pose is represented by 22 body joints. Then, each data sequence is down-sampled to 25 FPS. The data from S5 and S11 are used as test and validation datasets, while the remaining 5 subjects are used for training.

\subsubsection{CMU Mocap} CMU Mocap\footnote{http://mocap.cs.cmu.edu/. We only report results on 7 actions due to page limits. The results of "running" are presented in the \emph{supplementary material}.} is another well-known benchmark dataset. Each pose is represented as 25 body joints. The pre-processing procedure is the same as that of Human3.6M, and the dataset split in \cite{mao2019learning,dang2021msr} is adopted.

\subsubsection{3DPW} 3DPW \cite{marcard2018recovering} is a more challenging dataset and contains human motion sequences from indoor and outdoor scenarios. Like other baselines, a pose is represented as 23 body joints, and the official split is utilized \cite{mao2019learning,liu2020trajectorycnn,dang2021msr}.

\subsection{Comparison Settings}

\subsubsection{Metrics} The Mean Per Joint Position Error (MPJPE) is a standard evaluation metric used in previous studies \cite{martinez2017human,li2018convolutional,dang2021msr}. With the predicted motion sequence $\tilde{X}_{K+1:K+L}$ and corresponding ground truth $X_{K+1:K+L}$, the MPJPE loss is defined as
\begin{equation}
\mathcal{L}_{\text {MPJPE}}=\frac{1}{J \times L} \sum_{t=K+1}^{K+L} \sum_{j=1}^{J}\left\|\tilde{p}_{j, t}-p_{j, t}\right\|_{2},
\end{equation}
where $\tilde{p}_{j,t} \in \mathbb{R}^{D}$ is the predicted $j$-th joint position in frame $t$, while $p_{i, t} \in \mathbb{R}^{D}$ is the corresponding ground truth.

\subsubsection{Implementation Details}\label{sec:impl_detail} All experiments were conducted on one RTX 3080 Ti GPU with the Pytorch framework \cite{paszke2019pytorch}. For the state-of-the-art baseline models, their official implementations were adopted, and they were trained with default settings. For the comparison of DSTD-GCN and the baseline models for graph convolutions, PReLU \cite{he2015delving} was adopted as the activation function, and the channel number was set to 64. Meanwhile, all the models were trained with the Adam optimizer \cite{kingma2014adam} with an initial learning rate of $3e-3$, which was decayed by 0.9 for every 5 epochs; MPJPE was used as the loss function, and the batch size was set to 32; the reduction rate $r$ was set to 32 in the DSTD-GC. Besides, the test set and sequence length settings followed the experiment setting of Dang \textit{et al.} \cite{dang2021msr}. Moreover, the full test set was utilized, and the input length was set to 10, while the output length was set to 30 on 3DPW dataset and 25 on the other two datasets. The detailed model architecture is provided in the \emph{supplementary material}.

\begin{table}[!htp]
\setlength\tabcolsep{1.5pt}
\caption{Comparison of different graph convolution. ST-U and S-S are referred to as "spatiotemporal-unshared" and "sample-specific", respectively.}
\begin{center}
\label{tab:graph modeling}
\resizebox{0.48 \textwidth}{15mm}{
\begin{tabular}{cc|cc|ccccc|ccccccc}
\hline
\multirow{2}{*}{GCs} &
\multirow{2}{*}{Params.} &
\multicolumn{2}{c|}{Classifications} &
\multicolumn{5}{c|}{Constraints} &
\multicolumn{7}{c}{\textit{MPJPE}} \\
\multicolumn{1}{c}{} &
\multicolumn{1}{c|}{} &
ST-U &
S-S &
1  &
2  &
3  &
4  &
5  &
80 &
160 &
320 &
400 &
560 &
1000 &
Average \\
\hline
\hyperlink{stgc_intro}{ST} & 5.44M & - & - & & & & & &  8.70 & 16.16 & 31.34 & 39.16 & 53.91 & 85.30 & 39.09 \\
\hline
\hline
\hyperlink{vstdgc_intro}{VSTD} & 0.10M &\XSolidBrush & \XSolidBrush & \Checkmark & \Checkmark & \Checkmark & & & 8.48 & 16.38 & 33.63 & 42.33 & 58.37 & 93.05 & 42.04 \\
FC \cite{mao2019learning} & 0.29M &\XSolidBrush & \XSolidBrush & \Checkmark & \Checkmark & \Checkmark & & & 13.11 & 24.87 & 46.47 & 56.16 & 73.75 & 106.75 & 53.52 \\
\hline
STS \cite{sofianos2021space} & 0.45M &\Checkmark & \XSolidBrush & \Checkmark & & \Checkmark & \Checkmark & & 8.33 & 15.62 & 30.97 & 38.77 & 53.70 & 85.34 & 38.79 \\
\hyperlink{tsdgc_intro}{TSD} & 0.46M & \Checkmark & \XSolidBrush & \Checkmark & & \Checkmark & \Checkmark & & 7.85 & 14.97 & 30.61 & 38.94 & 54.90 & 87.51 & 39.13 \\
\hyperlink{stdgc_intro}{STD} & 0.46M & \Checkmark & \XSolidBrush & \Checkmark & & \Checkmark & \Checkmark & & 8.14 & 15.45 & 30.71 & 38.51 & 53.75 & 86.10 & 38.78 \\
\hline
\hyperlink{dtsdgc_intro}{DTSD} & 0.13M & \Checkmark & \Checkmark & \Checkmark & & & \Checkmark & \Checkmark & \underline{7.74} & \textbf{14.11} & \underline{29.72} & \underline{37.68} & \underline{52.47} & \underline{86.20} & \underline{37.99} \\
\hyperlink{dstdgc_intro}{DSTD} & 0.13M & \Checkmark & \Checkmark & \Checkmark & & & \Checkmark & \Checkmark & \textbf{7.36} & \underline{14.21} & \textbf{29.29} & \textbf{36.91} & \textbf{51.57} & \textbf{84.66} & \textbf{37.33} \\
\hline
\end{tabular}}
\end{center}
\end{table}
\vspace{-10pt}

\subsection{RQ1: Comparison with other GCs}\label{sec:gc_compare}

We compare the parameter number and prediction error of DSTD-GC against other graph convolutions in Table \ref{tab:graph modeling}. Specifically, three key aspects of DSTD-GCN are investigated: spatiotemporal-unshared decomposition, sample-specific correlation, and stacking order invariance. For a fair comparison, the prediction framework is kept unchanged, and only the GC in the Basic Conv unit is changed. Moreover, 
the linear transformation is changed to a two-layer $MLP$ for ST-GC, VST-GC, FC-GC, and STS-GC. For our DSTD-GC, the Basic Conv unit in Fig. \ref{fig:details} (a) is adopted, and one DS-GC is removed in the spatial modeling. Also, \hypertarget{tsdgc_intro}{TSD-GC} and \hypertarget{dtsdgc_intro}{DTSD-GC} are introduced by switching the order of spatial and temporal graph convolutions in STD-GC and DSTD-GC, respectively. For a fair comparison, we follow previous approaches \cite{mao2019learning,dang2021msr,sofianos2021space} and initialize adjacency matrices randomly for all graph convolutions.

\subsubsection{Spatiotemporal-shared GCs vs. Spatiotemporal-unshared GCs} \label{sec:sc_vs_uc} Since spatiotemporal-shared and spatiotemporal-unshared GCs are defined for decomposing ST-GC, they are compared with the ST-GC. It can be observed that spatiotemporal-shared GCs (STS, TSD, STD) have fewer parameters than spatiotemporal-unshared GCs but perform worse than ST-GC. Meanwhile, the spatiotemporal-unshared GCs obtain similar errors as ST-GC, which indicates that unshared spatial and temporal correlations are more suitable forms for describing spatiotemporal correlations because the unshared correlations can explicitly describe the varying spatiotemporal correlations in different body parts and different motion stages. Although spatiotemporal-shared GCs take fewer parameters than spatiotemporal-unshared ones, the corresponding models are not lightweight because these models need to stack more layers to achieve comparable performance. This will be shown in Sect. \ref{sec:efficiency_analysis}. Our DSTD-GC utilizes a shared vanilla adjacency matrix to store general correlations and a function to extract unshared correlation variances. Thus, our DSTD-GC is parameter-saving and explicitly models spatiotemporal correlations.

\subsubsection{Sample-generic GCs vs. Sample-specific GCs} We compare the two sample-generic GCs (\hyperlink{tsdgc_intro}{TSD-GC} and \hyperlink{stdgc_intro}{STD-GC}) with their sample-specific counterparts (DTSD-GC and DSTD-GC). It can be found that sample-specific GCs make consistent improvements in accuracy. This is because the learned sample-generic correlations cannot precisely depict the correlations in individual sequences and the correlation balance between various input samples. This global optimal strategy may introduce false correspondence that disturbs individual predictions. More investigations on this are shown in Sect. \ref{sec:vis_cdc}.

\subsubsection{Spatial-temporal Order vs. Temporal-spatial Order} As STD-GC and TSD-GC, DSTD-GC and DTSD-GC perform similarly, the modeling capability of spatiotemporal decomposed GC is invariant to the stacking order of spatial and temporal GCs. 

\begin{table*}[!ht]
\linespread{1}
\setlength\tabcolsep{1.5pt}
\begin{center}
\caption{Comparison of prediction results on the Human3.6M dataset. The best results are highlighted in \textbf{bold}, while the second best results are shown in \underline{underline}.}
\label{tab:h36m_all}
\resizebox{\textwidth}{56mm}{
\begin{tabular}{c|cccccc|cccccc|cccccc|cccccc}
\hline
Action & \multicolumn{6}{c|}{Walking} & \multicolumn{6}{c|}{Eating} & \multicolumn{6}{c|}{Smoking} & \multicolumn{6}{c}{Discussion} \\
\hline
Millisecond & 80 & 160 & 320 & 400 & 560 & 1000 & 80 & 160 & 320 & 400 & 560 & 1000 & 80 & 160 & 320 & 400 & 560 & 1000 & 80 & 160 & 320 & 400 & 560 & 1000 \\
\hline
Residual sup. \cite{martinez2017human} & 24.13 & 38.59 & 55.09 & 59.26 & 65.14 & 82.04 & 18.64 & 30.38 & 47.97 & 54.84 & 67.80 & 105.56 & 18.06 & 29.63 & 45.80 & 52.56 & 64.25 & 91.95 & 27.74 & 48.22 & 79.13 & 90.02 & 106.71 & 134.22 \\
DMGNN \cite{li2020dynamic}              & 15.52             & 27.42             & 46.20             & 54.96             & 58.86             & 83.74             & 10.34             & 19.84             & 37.60             & 46.66             & 57.95             & 86.55             & 9.94              & 18.90             & 35.02             & 42.82             & 53.23             & 77.76  & 14.77 & 30.86 & 62.05 & 75.26 &  93.24 & 123.67 \\
FC-GCN \cite{mao2019learning}           & 12.29             & 23.03             & 39.77             & 46.12             & 54.05             & \textbf{59.75} & 8.36              & 16.90             & 33.19             & 40.70             & 53.39             & 77.75             & 7.94              & 16.24             & 31.90             & 38.90             & 50.74             & 72.62 & 12.50 & 27.40 & 58.51 & 71.68 & 91.61 & 121.53 \\
Traj-CNN \cite{liu2020trajectorycnn}    & \underline{11.91} & \underline{22.54} & 38.66             & 45.71 & 54.49             & 62.01             &  8.41             & \underline{16.56} & \underline{32.44} & \underline{39.82} & 53.47             &  78.40            & 8.41              & 16.17             & \underline{31.06} & \underline{37.58} & \underline{49.30} & 72.31 & 11.74 & \underline{26.30} & 57.29 & 70.36 &  91.45 & 122.66 \\
STS-GCN \cite{sofianos2021space}      & 11.98             & 22.96             & 41.53             & 48.47             & 56.62             & 62.69             & \underline{7.90}  & 16.79             & 33.38             & 40.74             & 53.10             & \underline{76.73} & \underline{7.48}  & \underline{15.69} & 31.33             & 38.45             & 50.67             & 73.10 & \underline{11.39} & 26.41 & 57.78 & 71.08 &  91.16 & 120.79 \\
MSR-GCN \cite{dang2021msr}              & 12.16             & 22.65             & \textbf{38.64} & \underline{45.24}             & \underline{52.72} & 63.04             & 8.39              & 17.05             & 33.03             & 40.43             & \underline{52.54} & 77.11             & 8.02              & 16.27             & 31.32             & 38.15             & 49.45             & \underline{71.64} & 11.98 & 26.76 & \underline{57.08} & \underline{69.74} &  \underline{88.59} & \underline{117.59} \\
\hline
Ours                                  & \textbf{11.05} & \textbf{22.35} & \underline{38.81} & \textbf{45.19} & \textbf{52.70} & \underline{59.76} & \textbf{6.95} & \textbf{15.51} & \textbf{31.74} & \textbf{39.19} & \textbf{51.86} & \textbf{76.19} & \textbf{6.64} & \textbf{14.75} & \textbf{29.78} & \textbf{36.67} & \textbf{48.09} & \textbf{71.16}  & \textbf{9.98} & \textbf{24.37} & \textbf{54.53} & \textbf{67.40} & \textbf{87.00} & \textbf{116.30}  \\
\hline
\noalign{\smallskip}
\hline
Action & \multicolumn{6}{c|}{Directions} & \multicolumn{6}{c|}{Greetings} & \multicolumn{6}{c|}{Phoning} & \multicolumn{6}{c}{Posing} \\
\hline
Millisecond & 80 & 160 & 320 & 400 & 560 & 1000 & 80 & 160 & 320 & 400 & 560 & 1000 & 80 & 160 & 320 & 400 & 560 & 1000 & 80 & 160 & 320 & 400 & 560 & 1000 \\
\hline
Residual sup. \cite{martinez2017human}  & 21.93 & 37.38 & 61.51 & 71.86 & 88.09 & 122.08 & 35.28 & 61.98 & 99.06 & 111.04 & 127.44 & 160.59 & 21.05 & 35.49 & 57.89 & 67.26 & 83.74 & 130.34 & 32.99 & 59.71 & 104.87 & 123.67 & 158.74 & 300.18 \\
DMGNN \cite{li2020dynamic}              & 10.77 & 22.81 & 46.70 & 57.35 & 73.51 & 104.37 & 20.45 & 41.13 & 78.99 & 94.34 & 114.68 & 148.75 & 12.35 & 24.25 &  47.22 &  58.27 & 73.62 & 113.05 & 15.40 & 32.07 & 67.36 &  84.19 & 113.60 & 171.56 \\
FC-GCN \cite{mao2019learning}           &  8.97 & 19.87 & 43.35 & \underline{53.74} &  71.01 & 101.79 & 18.65 & 38.68 &  77.74 &  93.39 & 114.43 & 148.69 & 10.24 & 21.02 &  42.54 &  52.30 & 69.56 & 104.50 & 13.66 & 29.89 &  66.62 &  84.05 & 113.56 & 171.31 \\
Traj-CNN \cite{liu2020trajectorycnn}    &  8.69 & 19.29 & 43.57 & 54.36 &  74.56 & 109.42 & 15.81 & 35.12 &  73.56 &  \underline{88.89} & \underline{110.78} & 149.55 & 10.14 & 20.52 &  41.95 &  51.86 & 69.29 & 104.41 & 12.09 & \underline{26.94} &  \underline{62.44} &  \underline{79.33} & \underline{108.36} & 170.86  \\
STS-GCN \cite{sofianos2021space}        &  \underline{7.82} & \underline{18.72} & \underline{42.58} & 53.25 &  70.95 & 102.10 & \underline{15.33} & \underline{35.01} &  \underline{73.44} &  89.08 & 112.19 & \underline{143.91} & \underline{9.54} &  \underline{20.35} &  41.55 &  \underline{51.07} & 68.32 & \underline{103.69}  & \underline{11.61} & 27.60 &  63.85 &  81.23 & 111.68 & \underline{168.41}  \\
MSR-GCN \cite{dang2021msr}              &  8.61 & 19.65 & 43.28 & 53.82 &  71.18 & \underline{100.59} & 16.48 & 36.95 &  77.32 &  93.38 & 116.26 & 147.26 & 10.10 & 20.74 &  \underline{41.51} &  51.26 & \underline{68.29} & 104.27   & 12.79 & 29.38 &  66.95 &  85.01 & 116.27 & 174.33 \\
\hline
Ours                                    & \textbf{6.86} & \textbf{17.39} & \textbf{40.96} & \textbf{51.67} & \textbf{69.05} & \textbf{99.05} & \textbf{14.27} & \textbf{33.53} & \textbf{72.15} & \textbf{87.31} & \textbf{108.66} & \textbf{142.28} & \textbf{8.52} & \textbf{19.22} & \textbf{40.31} & \textbf{49.87} & \textbf{66.69} & \textbf{102.20}  & \textbf{10.08} & \textbf{25.40} & \textbf{60.60} & \textbf{77.34} & \textbf{106.54} & \textbf{163.31} \\
\hline
\noalign{\smallskip}
\hline
Action & \multicolumn{6}{c|}{Purchases} & \multicolumn{6}{c|}{Sitting} & \multicolumn{6}{c|}{Sitting Down} & \multicolumn{6}{c}{Taking Photo}  \\
\hline
Millisecond & 80 & 160 & 320 & 400 & 560 & 1000 & 80 & 160 & 320 & 400 & 560 & 1000 & 80 & 160 & 320 & 400 & 560 & 1000 & 80 & 160 & 320 & 400 & 560 & 1000 \\
\hline
Residual sup. \cite{martinez2017human}  & 30.54 & 52.89 & 86.31 & 99.07 & 121.00 & 169.66 & 22.08 & 38.58 & 64.59 & 75.94 & 96.33 & 145.90 & 27.73 & 47.24 & 79.55 & 93.33 & 117.92 & 170.19 & 21.53 & 37.30 & 63.83 & 75.78 & 95.70 & 145.82 \\
DMGNN \cite{li2020dynamic}              & 17.60 & 36.24 & 69.43 & 83.20 & 104.09 & 145.26 & 11.77 & 23.42 & 48.33 & 60.69 & 80.75 & 123.31 & 16.93 & 32.04 & 62.19 & 76.52 & 101.16 & 153.00 & 11.49 & 23.07 & 46.89 & 58.46 & 78.20 & 121.69 \\
FC-GCN \cite{mao2019learning}           & 15.60 & 32.78 & \underline{65.72} & \underline{79.25} & \underline{100.19} & 141.14 & 10.62 & 21.90 &  46.33 &  57.91 & 79.38 & 122.44 & 16.14 & 31.12 &  61.74 &  76.46 &  99.24 & 149.30 &  9.88 & 20.89 & 44.95 & 56.58 & 76.52 & 119.33 \\
Traj-CNN \cite{liu2020trajectorycnn}    & 14.54 & 31.88 & 66.55 & 80.75 & 103.65 & 141.01 & 10.97 & 21.17 &  45.48 &  57.50 & 78.95 & 120.12 & 16.13 & \underline{29.56} &  \underline{58.74} &  \underline{72.59} &  \underline{97.00} & \textbf{146.96} & 10.43 & 20.64 & 44.37 & 55.83 & 76.78 & 120.11 \\
STS-GCN \cite{sofianos2021space}        & \underline{13.87} & \underline{31.66} & 66.00 & 80.04 & 102.46 & 142.46 & \underline{9.63} & \underline{20.65} & \underline{45.22} & \underline{57.26} & 78.96 & 122.03 & \underline{14.98} & 29.60 & 59.41 & 73.55 & 98.80 & 149.52 & \underline{9.15} & \underline{19.87} & \underline{43.42} & \underline{54.99} & \underline{76.15} & \underline{118.76} \\
MSR-GCN \cite{dang2021msr}             & 14.75 & 32.39 & 66.13 & 79.64 & 101.63 & \underline{139.16} & 10.53 & 21.99 &  46.26 &  57.80 & \underline{78.20} & \underline{120.04} & 16.10 & 31.63 &  62.45 &  76.84 & 102.84 & 155.47 & 9.89 & 21.01 & 44.56 & 56.30 & 77.97 & 121.91\\
\hline
Ours                                    & \textbf{12.68} & \textbf{29.65} & \textbf{62.29} & \textbf{75.79} & \textbf{97.54} & \textbf{137.76} & \textbf{8.78} & \textbf{19.32} & \textbf{42.88} & \textbf{54.33} & \textbf{74.94} & \textbf{117.75}  & \textbf{14.10} & \textbf{28.03} & \textbf{57.33} & \textbf{71.18} & \textbf{96.08} & \underline{147.25} & \textbf{8.41} & \textbf{18.84} & \textbf{42.00} & \textbf{53.50} & \textbf{74.50} & \textbf{117.91} \\
\hline
\noalign{\smallskip}
\hline
Action & \multicolumn{6}{c|}{Waiting} & \multicolumn{6}{c|}{Walking Dog} & \multicolumn{6}{c|}{Walking Together} & \multicolumn{6}{c}{Average} \\
\hline
Millisecond & 80 & 160 & 320 & 400 & 560 & 1000 & 80 & 160 & 320 & 400 & 560 & 1000 & 80 & 160 & 320 & 400 & 560 & 1000 & 80 & 160 & 320 & 400 & 560 & 1000 \\
\hline
Residual sup. \cite{martinez2017human}  & 25.78 & 44.52 & 72.29 & 82.44 & 98.79 & 136.76 & 39.19 & 67.36 & 105.56 & 117.96 & 135.99 & 186.09 & 22.03 & 36.08 & 54.91 & 60.30 & 67.89 & 85.19 & 25.91 & 44.36 & 71.86 & 82.35 & 99.90 & 144.44 \\
DMGNN \cite{li2020dynamic}              & 12.89 & 25.70 &  51.27 &  62.84 & 79.82 & 112.99 & 26.09 & 50.47 & 88.90 & 102.55 & 118.44 & 156.09 & 13.76 & 25.55 & 44.49 & 53.47 & 59.41 & 79.06 & 14.67 & 28.91 & 55.51 & 67.44 & 84.04 & 129.06 \\
FC-GCN \cite{mao2019learning}           & 11.43 & 23.99 &  50.06 &  61.48 & 78.15 & 108.77 & 23.39 &  46.17 &  83.47 &  95.96 & \underline{110.98} & \textbf{146.24} & 10.47 & 21.04 & 38.47 & 45.19 & 54.71 & 66.96 & 12.68 & 26.06 & 52.29 & 63.58 & 81.17 & 114.14 \\
Traj-CNN \cite{liu2020trajectorycnn}    & 10.51 & \underline{21.76} &  \underline{45.79} &  \underline{56.29} & \underline{73.36} & \underline{104.53} & 21.30 &  43.29 &  80.77 &  94.50 & 115.56 & 153.50 & 10.30 &  21.11 & 38.48 & 44.82 & 54.78 & 68.00 & 12.09 & 24.86 & 50.74 & 61.95 & 80.78 & 114.92 \\
STS-GCN \cite{sofianos2021space}        & \underline{10.00} & 21.93 &  46.98 &  58.22 & 76.39 & 107.68 & 20.79 &  43.56 &  81.81 &  95.20 & 114.36 & 151.92 & \underline{10.06} & \underline{20.69} & 39.07 & 46.00 & 54.92 & \underline{62.91}  & \underline{11.43} & \underline{24.77} & \underline{51.16} & \underline{62.58} & \underline{81.12} & \underline{113.78} \\
MSR-GCN \cite{dang2021msr}              & 10.68 & 23.06 &  48.25 &  59.23 & 76.33 & 106.27 & 20.65 &  42.88 &  \underline{80.35} &  \underline{93.31} & 111.89 & 148.24 & 10.56 & 20.92 & \underline{37.40} & \underline{43.85} & \underline{52.94} & 65.94 & 12.11 & 25.56 & 51.64 & 62.93 & 81.14 & 114.19 \\
\hline
Ours                                    & \textbf{8.71} & \textbf{20.15} & \textbf{44.28} & \textbf{55.25} & \textbf{73.19} & \textbf{105.66}  & \textbf{19.64} & \textbf{41.82} & \textbf{77.61} & \textbf{90.24} & \textbf{109.84} & \underline{147.68} & \textbf{9.07} & \textbf{19.79} & \textbf{36.33} & \textbf{42.67} & \textbf{50.54} & \textbf{61.22} & \textbf{10.38} & \textbf{23.34} & \textbf{48.77} & \textbf{59.84} & \textbf{77.81} & \textbf{111.02} \\
\hline
\end{tabular}}
\end{center}
\end{table*}

\begin{table*}[!htp]
\setlength\tabcolsep{1.5pt}
\caption{Comparison of prediction results on the CMU Mocap dataset. The best results are highlighted in \textbf{bold}, while the second best results are shown in \underline{underline}.}
\centering
\label{tab:cmu_all}
\resizebox{\textwidth}{28mm}{
\begin{tabular}{c|cccccc|cccccc|cccccc|cccccc}
\hline
Action & \multicolumn{6}{c|}{Basketball} & \multicolumn{6}{c|}{Basketball Signal} & \multicolumn{6}{c|}{Directing Traffic} & \multicolumn{6}{c}{Jumping} \\
\hline
Millisecond & 80 & 160 & 320 & 400 & 560 & 1000 & 80 & 160 & 320 & 400 & 560 & 1000 & 80 & 160 & 320 & 400 & 560 & 1000 & 80 & 160 & 320 & 400 & 560 & 1000 \\
\hline
Residual sup. \cite{martinez2017human}  & 29.50 & 53.05 & 91.22 & 106.03 & 128.74 & 157.38 & 14.63 & 22.07 & 39.07 & 46.56 & 59.98 & 89.93 & 21.77 & 38.78 & 70.45 & 85.3 & 110.29 & 165.13 & 30.18 & 53.02 & 89.35 & 103.9 & 125.55 & 160.49 \\
DMGNN \cite{li2020dynamic}              & 14.97 & 27.07 & 49.36 & 61.45 & 84.83 & 145.18 &  4.91 &  8.78 & 15.93 & 19.60 & 27.43 & 47.28 & 10.62 & 19.63 & 36.03 & 44.34 & 62.87 & 115.71 & 19.47 & 36.77 & 67.42 &  81.28 & 105.62 & 148.53 \\
FC-GCN \cite{mao2019learning}           & 11.67 & 21.09 & 40.70 & 50.58 & 68.03 & 95.66 & 3.35 &  6.23 & 13.48 & 17.87 & 27.34 & 51.88 &  6.78 & 13.36 & 29.57 & 39.06 & 59.64 & 112.83 & 17.10 & 32.06 & 59.82 &  72.51 & 94.33 & 127.20 \\
Traj-CNN \cite{liu2020trajectorycnn}    & 11.84 & 19.12 & \underline{36.72} & 46.02 & 62.47 & 95.76 &  4.42 &  6.20 & 12.29 & 16.19 & 25.48 & 51.76 &  6.95 & \underline{11.03} & \underline{25.89} & 40.99 & 54.76 & 112.43 & \underline{14.88} & \underline{27.01} & \underline{55.31} &  71.72 & 94.23 & 126.97 \\
STS-GCN \cite{sofianos2021space}        & \underline{10.23} & \underline{18.67} & 36.93 & \underline{45.98} & \underline{61.19} & 91.36 &  \underline{2.96} & \underline{5.52} & \underline{12.12} & \underline{16.12} & \underline{25.15} & 50.88 & \underline{5.95} & 11.99 & 27.55 & \underline{36.75} & 57.05 & \underline{111.53} & 15.66 & 30.63 & 59.13 & 71.87 & 93.32 & \textbf{125.94}\\
MSR-GCN \cite{dang2021msr}              & 10.28 & 18.94 & 37.68 & 47.03 & 62.01 & \textbf{86.27} &  3.04 &  5.63 & 12.51 & 16.61 & 25.46 & \underline{50.04} &  6.13 & 12.61 & 29.39 & 39.24 & \underline{50.49} & 114.58 & 15.19 & 28.86 & 55.98 &  \underline{69.12} & \underline{92.40} & 126.18\\
\hline
Ours                                     & \textbf{9.60} & \textbf{17.64} & \textbf{35.44} & \textbf{44.43} & \textbf{59.97} & \underline{88.44} & \textbf{2.57} & \textbf{4.72} & \textbf{10.37} & \textbf{13.86} & \textbf{21.85} & \textbf{46.17} & \textbf{5.02} & \textbf{10.01} & \textbf{23.35} & \textbf{31.40} & \textbf{49.28} & \textbf{99.57} & \textbf{12.81} & \textbf{26.05} & \textbf{54.62} & \textbf{68.47} & \textbf{91.83} & \underline{126.07}  \\
\hline
\noalign{\smallskip}
\hline
Action & \multicolumn{6}{c|}{Soccer} & \multicolumn{6}{c|}{Walking} & \multicolumn{6}{c|}{Wash Window} & \multicolumn{6}{c}{Average} \\
\hline
Millisecond & 80 & 160 & 320 & 400 & 560 & 1000 & 80 & 160 & 320 & 400 & 560 & 1000 & 80 & 160 & 320 & 400 & 560 & 1000 & 80 & 160 & 320 & 400 & 560 & 1000 \\
\hline
Residual sup. \cite{martinez2017human}   & 26.51 & 46.98 & 81.45 & 96.18 & 117.9 & 139.06 & 14.61 & 22.87 & 36.09 & 40.90 & 51.10  & 69.49 & 19.32 & 31.77 & 56.05 & 66.00 & 83.62 & 125.87 & 22.36 & 38.36 & 66.24 & 77.84 & 96.74 & 129.62 \\
DMGNN \cite{li2020dynamic}               & 17.64 & 31.86 & 56.81 & 68.84 & 92.70 & 130.80 & 12.23 & 21.89 &  36.03 &  41.32 & 51.46 & 64.68 & 9.10 & 16.90 & 32.55 &  41.13 & 57.09 & 97.08 & 12.70 & 23.27 & 42.01 & 51.13 & 68.86 & 107.04 \\
FC-GCN \cite{mao2019learning}           & 13.62 & 24.30 & 44.40 & 54.31 & 73.14 & 111.64 &  6.74 & 11.09 &  18.08 &  20.95 & 25.16 & \textbf{32.38} &  5.87 & 11.33 & 24.14 &  30.95 & 43.44 & \textbf{66.93} & 9.30 & 17.06 & 32.89 & 40.89 & 55.86 & 85.50 \\
Traj-CNN \cite{liu2020trajectorycnn}    & 13.46 & 21.25 & 38.65 & 47.26 & \underline{62.66} & \underline{97.33} &  7.69 & 11.28 &  18.02 &  20.62 & 25.67 & 40.35 & 6.64 & 11.04 & 24.14 & 31.22 & 44.19 & 71.34 &  9.41 & 15.27 & 30.15 & 39.14 & 52.63 & 85.13 \\
STS-GCN \cite{sofianos2021space}        & 11.30 & 20.45 & 39.04 & 48.88 & 69.12 & 102.54 & 6.87 & 11.29 & 18.13 & 21.06 & 26.12 & 37.86 & 5.44 & \underline{10.84} & 23.90 & \underline{30.72} & \underline{44.00} & 71.42 &  8.33 & 15.62 & 30.97 & 38.77 & 53.70 & 84.50 \\
MSR-GCN \cite{dang2021msr}              & \underline{10.92} & \underline{19.39} & 37.41 & \underline{47.01} & 65.26 & 101.86 & \underline{6.39} & \textbf{10.25} & \underline{16.89} & \underline{20.05} & \underline{25.49} & 36.82  &  \underline{5.41} & 10.94 & 24.51 & 31.80 & 45.14 & 70.19 &  \underline{8.19} & \underline{15.20} & \underline{30.53} & \underline{38.65} & \underline{52.32} & \underline{83.70} \\
\hline
Ours                                    & \textbf{10.25} & \textbf{18.96} & \textbf{36.79} & \textbf{45.65} & \textbf{62.29} & \textbf{96.93} & \textbf{6.34} & \underline{10.35} & \textbf{16.09} & \textbf{18.62} & \textbf{23.28} & \underline{33.56} & \textbf{4.75} & \textbf{9.53} & \textbf{21.98} & \textbf{28.99} & \textbf{42.48} & \underline{68.93} &  \textbf{7.33} & \textbf{13.90} & \textbf{28.37} & \textbf{35.91} & \textbf{50.11} & \textbf{79.95} \\
\hline
\end{tabular}}
\end{table*}

\subsection{RQ2: Comparison with the State-of-the-art}

To evaluate the performance of DSTD-GCN, the quantitative and qualitative results on Human3.6M, CMU Mocap, and 3DPW datasets are presented. First, quantitative results are provided by comparing DSTD-GCN with state-of-the-art methods, and then qualitative results are provided. Following previous studies \cite{mao2019learning,dang2021msr}, the results are divided into short-term ($<500$ ms) and long-term ($>500$ ms) predictions.

\subsubsection{Baselines} \label{sec:baseline_intro} We select six state-of-the-art baselines, including the RNN-based model (Residual sup \cite{martinez2017human}), CNN-based model (Traj-CNN \cite{liu2020trajectorycnn}), and GCN-based model (FC-GCN \cite{mao2019learning}, DMGNN \cite{li2020dynamic} and MSR-GCN \cite{dang2021msr} and STS-GCN \cite{sofianos2021space}). For GCN-based models, FC-GCN, DMGNN, and MSR-GCN belong to the sample-generic spatiotemporal-shared method, and STS-GCN\footnote{The model from RQ1 was utilized. A comparison between our implementation and the official one is presented in the \emph{supplementary material}.} is the only sample-generic spatiotemporal-unshared method. For the graph correlations, DMGNN adopts fixed correlations from body connections, while the other three GCN-based methods initialize the adjacency matrix randomly and then optimize it through back-propagation. Among all the baseline methods, FC-GCN, Traj-CNN, MSR-GCN, STS-GCN, and our approach utilize the prediction framework demonstrated in Fig. \ref{fig:details} (a).

\begin{figure}[!ht]
  \centering
  \vspace{-0.2cm}
  \includegraphics[width=0.48\textwidth]{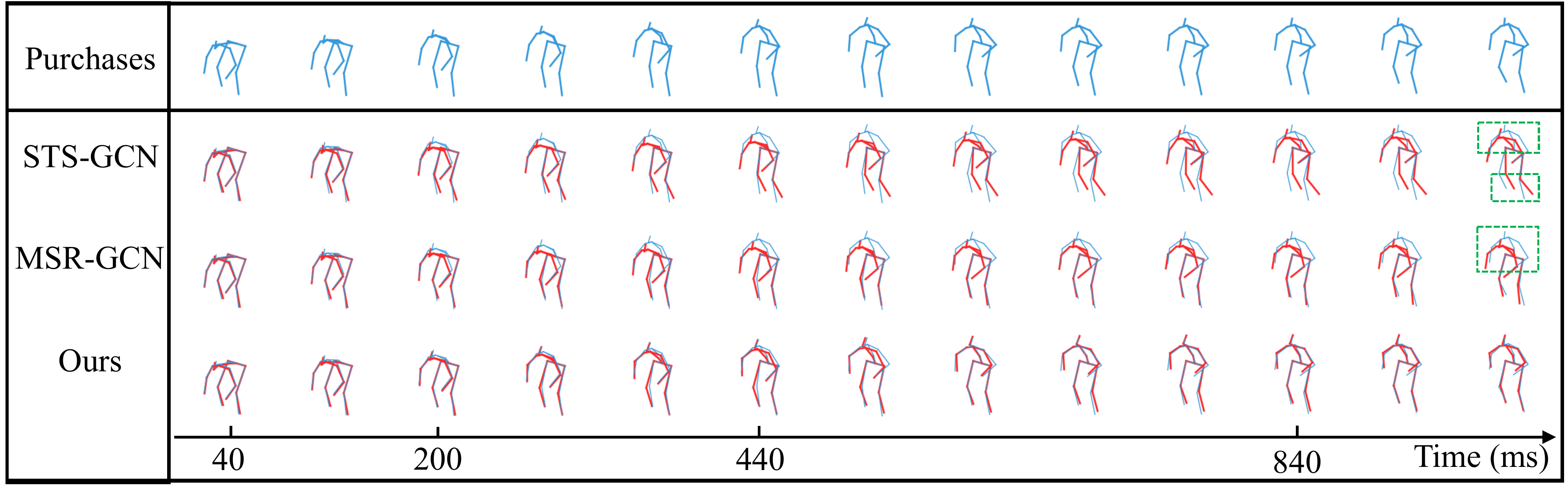}
  \vspace{-0.2cm}
  \caption{Visualization of predicted samples of state-of-the-art methods on the Human3.6M dataset. The green boxes show our improvements.}
  \label{fig:skeleton_comparison_h36m}
  \vspace{-0.5cm}
\end{figure}

\begin{figure}[!ht]
  \centering
  \includegraphics[width=0.48\textwidth]{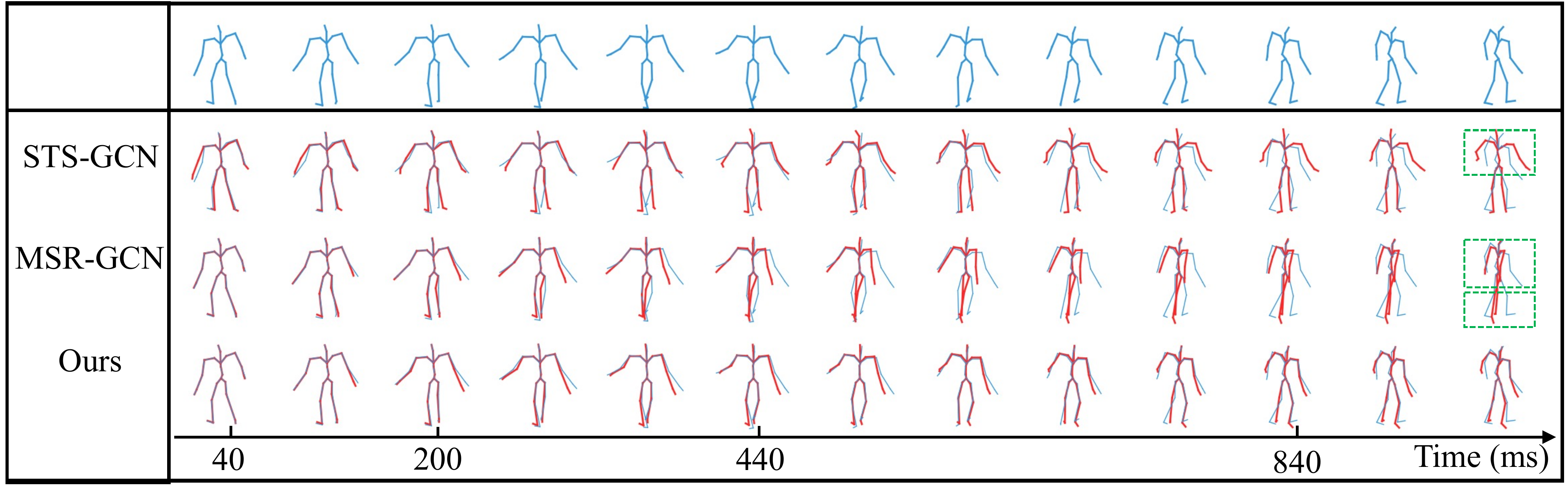}
  \vspace{-0.2cm}
  \caption{Visualization of predicted samples of state-of-the-art methods on the 3DPW dataset. The green boxes show our improvements.}
  \label{fig:skeleton_comparison_3dpw}
  \vspace{-0.2cm}
\end{figure}

\subsubsection{Results} 
\begin{figure*}[htbp]
  \centering
  \includegraphics[width=\textwidth]{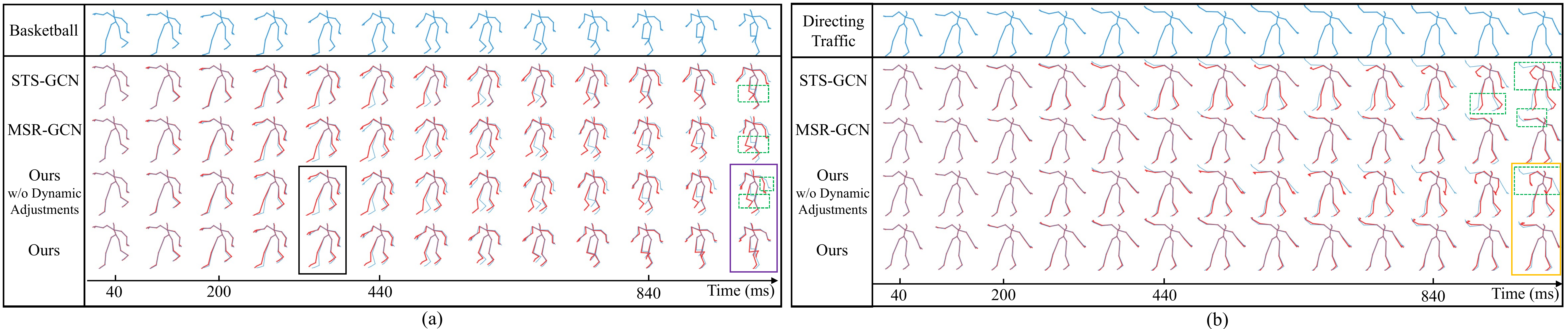}
  \vspace{-0.8cm}
  \caption{Visualization of the predicted results of the state-of-the-art methods on two action examples on the CMU Mocap dataset. The blue and red poses indicate ground truths and predictions, respectively. (a) Basketball. (b) Directing Traffic.}
  \label{fig:skeleton_comparison}
  \vspace{-0.3cm}
\end{figure*}

The detailed quantitative comparisons for both short-term and long-term prediction results are presented in Table \ref{tab:h36m_all} and Table \ref{tab:cmu_all}. Meanwhile, we also present the average results of selected frames in Table \ref{tab:compare_summary} and Table \ref{tab:compare_summary_3dpw}. Apparently, methods with the prediction framework make more accurate predictions with fewer parameters. Meanwhile, GCN-based methods generally outperform the RNN-based method (Residual Sup.) and CNN-based methods (Traj-CNN). For GCN-based methods, spatiotemporal-unshared methods outperform spatiotemporal-shared methods in terms of prediction accuracy and parameter numbers. Note that MSR-GCN contains 68 GC layers and combines DCT temporal encoding \cite{mao2019learning} with multi-scale modeling, while STS-GCN and DSTD-GCN only contain 7 GC layers with raw 3D inputs. With constrained dynamic correlation modeling, DSTD-GCN also models sample-specific correlations and achieves the best prediction accuracy with the fewest parameters. Even when the motion sequences become more challenging in the 3DPW dataset, our model still makes consistent improvements over the other approaches. For qualitative comparison, our method is compared with two state-of-the-art approaches on three benchmark datasets, and the examples are shown as Fig. \ref{fig:skeleton_comparison_h36m}, \ref{fig:skeleton_comparison_3dpw}, and  \ref{fig:skeleton_comparison}. In each of these examples, our method makes more accurate predictions. Specifically, we analyze the examples from the CMU Mocap dataset in detail. In the basketball scenario, spatiotemporal-unshared methods generally work better than spatiotemporal-shared ones because spatiotemporal-unshared methods explicitly model individual joints' spatiotemporal correlations and thus can infer a large movement from tiny motion cues. Here, both STS-GCN and our method can capture the backward trend of the right leg. However, sample-generic modeling of spatiotemporal correlations may disturb individual motion predictions, especially in some static action scenarios. As shown in the directing traffic scenario, most joints stay still, but STS-GCN amplifies the motion of the right arm and infers leg movement mistakenly. With constrained dynamic correlation modeling, our method can dynamically model sample-specific motion patterns and thus make more accurate predictions under different scenarios. Due to page limits, we present more visual comparison results in the \emph{supplementary material}.

\begin{table}[!ht]
\setlength\tabcolsep{1pt}
\centering
\caption{Comparison summary of the average MPJPE, parameter numbers, and inference time per iteration on Human3.6M and CMU Mocap datasets.}
\label{tab:compare_summary}
\begin{tabular}{c|ccc|ccc}
\hline
\multirow{2}{*}{Model} & \multicolumn{3}{c}{Human3.6M} & \multicolumn{3}{|c}{CMU Mocap} \\
\cline{2-7}
& MPJPE & Params. & Time & MPJPE & Params. & Time \\
\hline
DMGNN \cite{li2020dynamic}   & 63.27 & 46.90M & 422ms & 50.84 & 46.94M & 590ms \\
FC-GCN \cite{mao2019learning}  & 58.32 &  2.55M & 46ms & 40.30 & 2.70M & 47ms \\
Traj-CNN \cite{liu2020trajectorycnn} & 57.56 &  1.20M & 95ms & 38.62 & 1.20M & 110ms \\
STS-GCN \cite{sofianos2021space} & 57.47 &  0.40M & \textbf{39ms} & 38.64 & 0.46M & \textbf{41ms} \\
MSR-GCN \cite{dang2021msr} & 57.88 &  6.30M & 70ms & 38.10 & 6.37M & 82ms \\
\hline
Ours     & \textbf{55.19} &  \textbf{0.18M} & 48ms & \textbf{35.93} & \textbf{0.20M} & 50ms \\
\hline
\end{tabular}
\end{table}

\begin{table}[!ht]
\setlength\tabcolsep{1.5pt}
\centering
\caption{Comparison summary of MPJPE on 3DPW dataset}
\label{tab:compare_summary_3dpw}
\begin{tabular}{c|ccccc|c|c|c}
\hline
\multirow{2}{*}{Model} & \multicolumn{6}{c|}{MPJPE} & \multirow{2}{*}{Params.} & \multirow{2}{*}{Time} \\
\cline{2-7}
& 200 & 400 & 600 & 800 & 1000 & Average & \\
\hline
DMGNN \cite{li2020dynamic}   & 37.3 & 67.8 & 94.5 & 109.7 & 123.6 & 86.6 & 46.91M & 470ms \\
FC-GCN \cite{mao2019learning}  & 35.6 & 67.8 & 90.6 & 106.9 & 117.8 & 83.7 & 2.60M & 46ms \\
Traj-CNN \cite{liu2020trajectorycnn} & 30.0 & 59.7 & 85.3 & 99.1 & 107.7 & 76.4 & 1.20M & 103ms \\
STS-GCN \cite{sofianos2021space} & 27.2 & 52.6 & 75.1 & 94.1 & 107.9 & 71.4 & 0.49M & \textbf{41ms} \\
MSR-GCN \cite{dang2021msr} & 37.4 & 65.4 & 83.1 & 96.3 & 106.2 & 77.7 & 6.31M & 73ms \\
\hline
Ours     & \textbf{24.5} & \textbf{49.9} & \textbf{69.9} & \textbf{85.2} & \textbf{96.3} & \textbf{65.2} & \textbf{0.19M} & 49ms \\
\hline
\end{tabular}
\end{table}

\begin{figure*}[htbp]
  \centering 
  \includegraphics[width=1\textwidth]{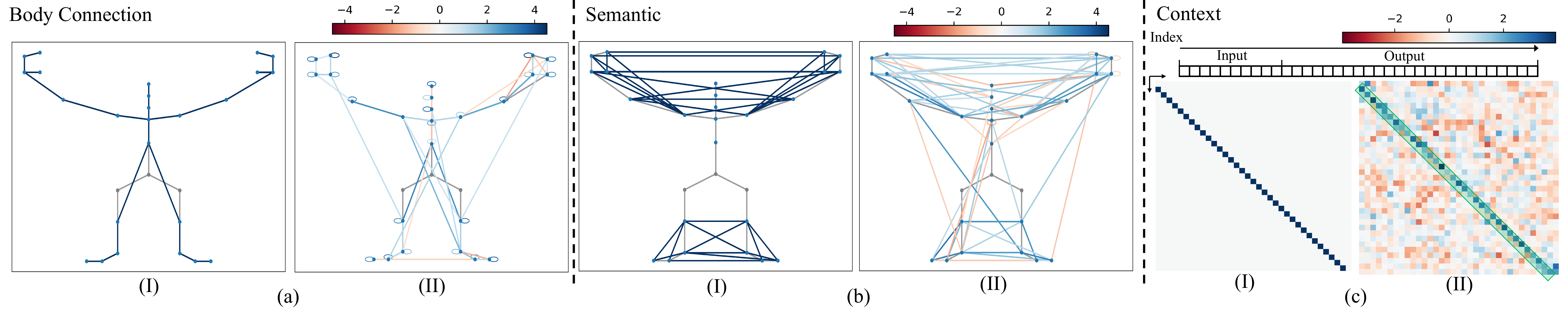}
  \vspace{-0.8cm}
  \caption{Visualization of constrained correlation. (a) and (b) indicate spatial correlations and (c) indicates temporal correlations. (I) is the initialized mask, where blue colors indicate predefined connections. For spatial correlation, we also show the filtered points and their natural connections in black. (II) are optimized constrained correlations, where color brightness indicates connection strength.}
  \label{fig:constrained_correlation}
  \vspace{-0.4cm}
\end{figure*}

\begin{figure}[!ht]
  \centering
  \includegraphics[width=0.48\textwidth]{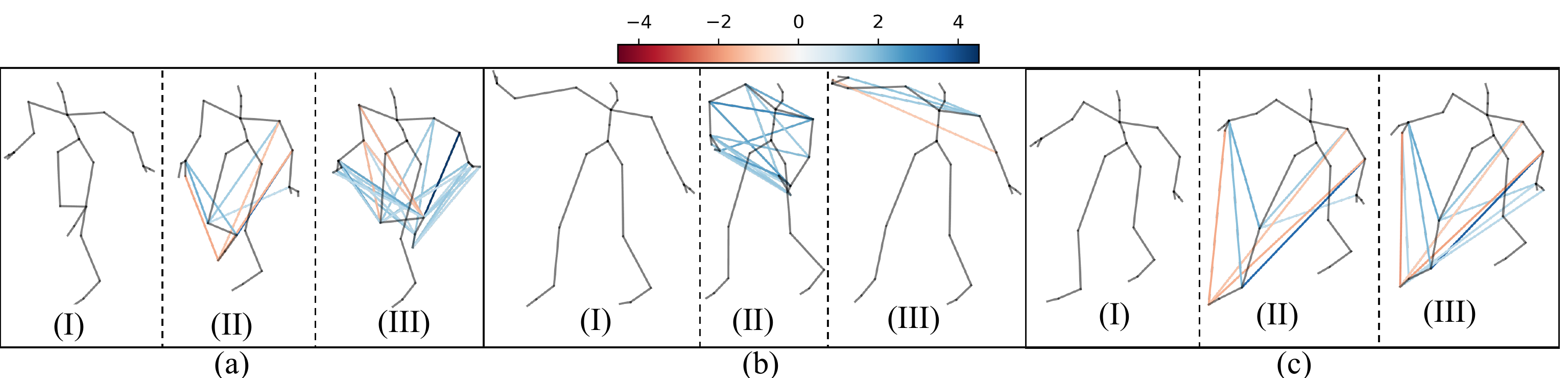}
  \vspace{-0.2cm}
  \caption{Visualization of spatial correlation adjustments. (a), (b) and (c) are three sample poses (black, orange, and purple boxes in Fig. \ref{fig:skeleton_comparison}). (I) , (II), and (III) are the ground truth pose, the predicted pose with constrained correlation, and the prediction with constrained dynamic correlation. For (II) and (III), we show certain pairs of constrained dynamic correlations for clarity.}
  \label{fig:spatial_adjustment}
  \vspace{-0.6cm}
\end{figure}

\begin{figure}[!ht]
  \centering
  \includegraphics[width=0.48\textwidth]{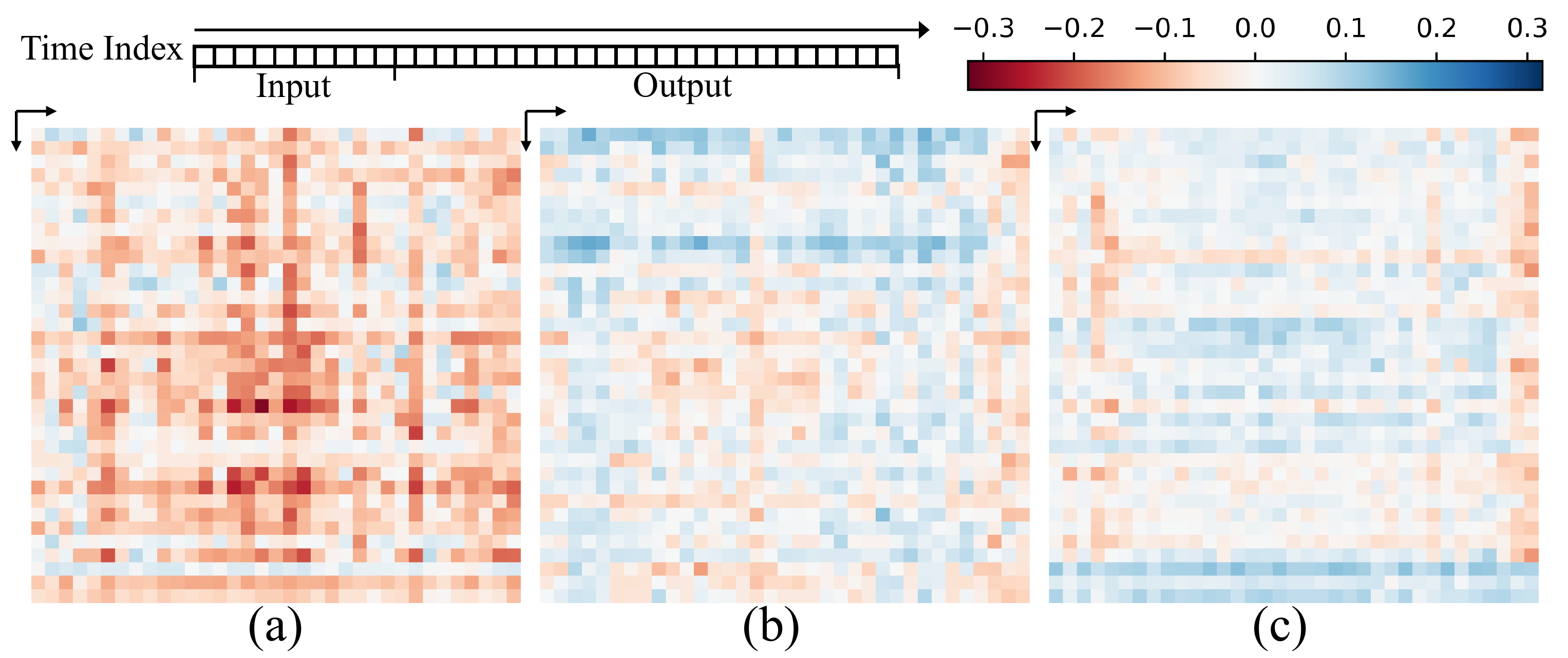}
  \vspace{-0.2cm}
  \caption{Visualization of temporal correlation adjustments. (a) and (b) are from the right foot, while (c) is from the head. (a) and (c) are from "Basketball", while (b) is from "Directing Traffic". Color brightness indicates connection strength.}
  \label{fig:temporal_adjustment}
  \vspace{-0.8cm}
\end{figure}

\subsubsection{Visualization of Constrained Dynamic Correlations}\label{sec:vis_cdc} In this section, we analyze different aspects of constrained dynamic correlation. We first visualize and evaluate the constrained correlation in Fig. \ref{fig:constrained_correlation}. Specifically, we present the spatial correlation as a frontally posing man and the temporal correlation as a vanilla matrix. For clarity, we represent the three strongest connections for each joint. We find that most optimized connections follow the predefined masks, which demonstrates the importance of the proposed prior knowledge initialization. Besides, some strong connections outside predefined masks intuitively contribute to accurate motion prediction, such as the hands to the legs and the torso and long-term frame links.

After the constrained correlation analysis, we use two example sequences in Fig. \ref{fig:skeleton_comparison} to better understand why dynamic adjustments take effect. First, we remove dynamic adjustments in DSTD-GC and compare the visualization sequences (Ours w/o Dynamic Adjustments) with the results from the full model. Then, we visualize the sample spatial and temporal adjustments from the two motion sequences in Fig. \ref{fig:spatial_adjustment} and Fig. \ref{fig:temporal_adjustment}. Combining the example comparison with the correlation visualization, we obtain two observations:
(1) Sample-specific correspondence adjustments contribute to accurate motion prediction. As shown in Fig. \ref{fig:spatial_adjustment} (a), the adjustment strengths from the arms to the right leg are enhanced in "Basketball". With these enhancement cues, the model predicts more accurate leg movement. Similarly, the connection strengths between the two arms are reduced in "Directing Traffic" in Fig. \ref{fig:spatial_adjustment} (b). Without these adjustments, the constrained correlation tends to infer wrong correspondences between the arms and amplify their movement mistakenly. Moreover, we also present temporal correspondence adjustments from the right ankle joint in Fig. \ref{fig:temporal_adjustment} (a) and (b), where we find the adjustment strengths are generally stronger in the basketball case. As the right foot moves backward in the basketball example, it correlates to other vertices of the trajectory more intensively. (2) Spatiotemporal-unshared correspondence adjustments reflect vital correlation variations in different movement stages/joints. For spatial modeling, the left arm moves in the late stage of the “Basketball” sequence. As shown in Fig. \ref{fig:temporal_adjustment} (a) and (c), the connection strengths between the left arm and the right leg are larger than those in the early movement stages. For temporal modeling, the knee joint moves continuously, so its temporal correlations are stronger than the ones of the head joint ((a) and (c) in Fig. \ref{fig:temporal_adjustment}).
Although temporal adjustment strengths are much smaller than constrained temporal correlation, they introduce critical correlation refinements for accurate motion prediction. As shown in Table \ref{tab:adjustments}, the performance drops sharply when we retrain the model without temporal adjustments.

\begin{table}[!ht]
\vspace{-0.2cm}
\setlength\tabcolsep{1.2pt}
\centering
\caption{Impact of Dynamic Temporal Correlation Adjustments}
\label{tab:adjustments}
\begin{tabular}{c|cccccc|c}
\hline
\multirow{2}{*}{Model Variant} & \multicolumn{7}{c}{MPJPE} \\
\cline{2-8}
& 80 & 160 & 320 & 400 & 560 & 1000 & Average \\
\hline
Without temporal adjustments  & 7.67 & 14.78 & 29.73 & 37.42 & 52.93 & 83.97 & 37.75 \\
Full model  & \textbf{7.33} & \textbf{13.90} & \textbf{28.37} & \textbf{35.91} & \textbf{50.11} & \textbf{79.95} & \textbf{35.93} \\
\hline
\end{tabular}
\end{table}

\vspace{-0.6cm}

\begin{table}[!ht]
\setlength\tabcolsep{1pt}
\centering
\caption{Comparison of Convolution for Basic Conv unit, Number of Basic Conv units in a basic block and basic block Number for Model with Prediction Framework.}
\label{tab:compare_prediction_framework}
\begin{tabular}{c|c|c|c}
\hline
Model & Convolution & Conv Number & Block Number  \\
\hline
Traj-CNN \cite{liu2020trajectorycnn} & CNN & 7 & 4 \\
\hline
FC-GCN \cite{mao2019learning}  & FC-GC \cite{mao2019learning} & 2 & 12  \\
MSR-GCN \cite{dang2021msr} & FC-GC \cite{mao2019learning} & 2 & 33 \\
STS-GCN \cite{sofianos2021space} & STS-GC \cite{sofianos2021space} & 1 & 5  \\
\hline
Ours  & \hyperlink{dstdgc_intro}{DSTD-GC} & 1 & 5   \\
\hline
\end{tabular}
\end{table}

\subsubsection{Effectiveness Analysis} \label{sec:efficiency_analysis} The model parameters and inference time of a single iteration are presented in Table \ref{tab:compare_summary} and Table \ref{tab:compare_summary_3dpw}. Generally, spatiotemporal-unshared methods outperform other methods in prediction accuracy and parameter number. For the models with the prediction framework, we also report the basic block number and number of convolutions in all basic blocks in Table \ref{tab:compare_prediction_framework}. Although the spatiotemporal-shared GC has fewer parameters than spatiotemporal-unshared GC as a single convolution unit, it needs to stack more layers to achieve comparable performance. The overall stacking overhead outweighs their lightweight benefits at an individual level. Meanwhile, our method outperforms other methods in prediction accuracy with the fewest parameters. The parameter number of our method is half of the most lightweight model STS-GCN and only $3\%$ of the best baseline MSR-GCN on the CMU Mocap dataset. Besides, our method is slightly slower than STS-GCN due to the additional computation of correspondence adjustments.

\begin{table}[!ht]
\setlength\tabcolsep{1.2pt}
\centering
\caption{Comparison of MPJPE for Ablation Variants.}
\label{tab:ablation}
\resizebox{0.48\textwidth}{13mm}{
\begin{tabular}{c|c|cccccc|c}
\hline
\multirow{2}{*}{ID} & \multirow{2}{*}{Model Variant} & \multicolumn{7}{c}{MPJPE} \\
\cline{3-9}
& & 80 & 160 & 320 & 400 & 560 & 1000 & Average \\
\hline
A & Only constrained correlation & 7.84 & 15.02 & 30.81 & 38.90 & 54.26 & 88.74 & 39.26 \\
B & Only dynamic correlation & 10.68 & 19.34 & 35.69 & 43.53 & 57.56 & 89.81 & 42.77 \\
C & Reversing feature adjustment order & 8.62 & 16.37 & 33.15 & 41.90 & 58.55 & 90.69 & 41.55 \\
D & Without prior initialization & 7.58 & 14.25 & 29.02 & 36.73 & 51.35 & 81.81 & 36.79 \\
E & Replacing DS-GC, DT-GC by S-GC, T-GC & 7.99 & 14.97 & 30.23 & 37.53 & 52.96 & 84.23 & 37.99 \\ 
F & Replacing DT-GC by DS-GC & 8.89 & 16.24 & 32.48 & 40.87 & 56.85 & 91.38 & 41.12 \\
G & Replacing DS-GC by DT-GC & 7.68 & 15.07 & 31.47 & 39.78 & 55.13 & 87.14 & 39.38 \\
\hline
H &  Full model  & \textbf{7.33} & \textbf{13.90} & \textbf{28.37} & \textbf{35.91} & \textbf{50.11} & \textbf{79.95} & \textbf{35.93} \\
\hline
\end{tabular}}
\end{table}

\subsection{RQ3: Ablation Study}

In this section, some major components of our method are investigated by comparing DSTD-GCN to its variants and other state-of-the-art methods.

\subsubsection{Constrained Dynamic Correlation Modeling}
First, the performance of the constrained and dynamic correlation is investigated by only using the constrained vanilla adjacency matrix and dynamic adjacency matrix for feature aggregation, respectively. The results are shown as A and B in Table \ref{tab:ablation}, and two findings can be obtained by comparing the average errors: (1) Constrained correlations cannot express erratic patterns in human motions, and there should be an adjustment function to complete the vertex correspondences for each motion sequence, as H outperforms A by 3.33. (2) With respect to the constrained dynamic correlation characteristic, explicit separate modeling is better than implicit united modeling, as variant B extracts both correlations simultaneously and performs worse than H by 6.84. Interestingly, A outperforms B by 3.51. Since the constrained correspondences are presented across all samples, and the dynamic correspondences are presented for each individual, these correlations are optimized toward two distinct directions, and neither of them can reach the optimum under the united optimization scenario. Meanwhile, the updating order is investigated by using the constrained feature to adjust the dynamic feature (C). C performs worse than A by 2.29, which indicates that dynamic correlations adjust constrained correlations but not vice versa.

\subsubsection{Prior Constrained Adjacency} The improvement in average MPJPE is compared for prior constrained adjacency and constrained dynamic correlation modeling. To show the effectiveness of the prior constrained adjacency, we designed a model with random constrained adjacency initialization (D). For constrained dynamic correlation modeling, DS-GC and DT-GC are replaced by S-GC and T-GC, respectively (E). As shown in Table \ref{tab:ablation}, the reduction of prior connections (0.86) is much less than the reduction of constrained dynamic correlation modeling (2.06). Besides, other state-of-the-art baselines do not obtain prior connections, and D still outperforms the best baseline MSR-GCN by 1.31 in average MPJPE, indicating the effectiveness of our proposed constrained dynamic correlation modeling.

\subsubsection{Spatial and Temporal GCs} To investigate the contribution of spatial and temporal GCs, the spatial and temporal graph convolutions are replaced with their spatiotemporal-equivalent variants in the full model. As shown as F and G in Table \ref{tab:ablation}, we find that (1) The full model (H) outperforms F and G in average MPJPE by 5.19 and 3.45, respectively, indicating that individual spatial or temporal GCs cannot make accurate predictions and emphasizing the importance of spatiotemporal correlation modeling; (2) H outperforms G by 1.74, demonstrating that temporal correlations are more important than spatial correlations for human motion prediction.

\section{Discussions}

\subsection{Limitations} We only design a multi-layer perceptron based on pair-wise concatenation to extract frame-wise/joint-wise spatial/temporal correlation adjustments. This exhaustive correlation modeling strategy calculates correlations between some unrelated vertices. A more powerful and efficient correlation modeling strategy can be designed. Another limitation is the scalability of the prediction framework, where input and output sequence length is fixed. This limitation hinders us from directly adopting the model to motion sequences with arbitrary lengths. This is a common problem for several state-of-the-art models \cite{mao2019learning,cui2020learning,dang2021msr,liu2020trajectorycnn}. 

\subsection{Broader Impacts}

Apart from human motion prediction, our proposed DSTD-GCN has two broader impacts.

First, spatiotemporal graphs are presented in a wide range of prediction applications, including point clouds \cite{mersch2021selfsupervised, gomes2021spatiotemporal}, traffic flows \cite{yu2017spatio, yang2016optimized},  trajectory \cite{yang2021novel, shi2021sgcn}, etc. \cite{kong2021spatiotemporal,yu2021deepsg}, where spatiotemporal-unshared decomposition and sample-specific correlation are crucial for feature representation. DSTD-GC can be used as a powerful basic convolution unit for model design in the above applications.

Second, the key idea of constrained dynamic correlation modeling is to decouple graph correlations into common patterns presented in all samples and unique patterns presented in each instance. The sample-generic and sample-specific patterns are intuitively optimized in distinct directions and should be modeled differently. Explicitly modeling these two patterns with constrained dynamic correlation modeling can reduce the optimization difficulty and might inspire adjacency modeling for other graph-based visual learning tasks \cite{liu2021light, liu2021instance, zhang2020weakly, liu2020comprehensive, shi2021sgcn}. In many of these works, the graph correlations are set either as model parameters, which cannot model sample-specific graph correlations or as attention weights from input features, where sample-generic correlations are ignored. Their graph convolution representation abilities can potentially be enhanced if they can separately model sample-generic and sample-specific correlations.

For example, our method can potentially be adopted in trajectory prediction. It is obvious that cars move in similar patterns, which are different from the ones from the pedestrians. Thus, the sample-generic correlation can be utilized here to represent inherent trajectory pattern similarities within cars/pedestrians. On the other hand, a car's movement pattern depends on surrounding cars and pedestrians under different circumstances, which could be depicted by the sample-specific correlation adjustments. Due to the page limits, we cannot conduct experiments to show this. We plan to test the effectiveness of DSTD-GC in future work.

\section{Conclusions}

In this work, a novel Dynamic SpatioTemporal Decompose Graph Convolutions (DSTD-GC) is proposed for human motion prediction. DSTD-GC employs constrained dynamic correlation modeling, which extends the conventional graph convolution by combining constrained correlations (from training or prior knowledge) and dynamic correlations from input motion sequences. Mathematical analyses and extensive experiments are conducted to illustrate the powerful spatiotemporal modeling capability of DSTD-GC. The results indicate that DSTD-GC breaks certain constraints of state-of-the-art graph convolutions on spatiotemporal graphs. With this strategy, we propose DSTD-GCN, which outperforms other state-of-the-art methods in prediction accuracy with the fewest parameters.


\appendix
\section*{A. Formula Derivation From DSTD-GC to the Sample-specific Spatiotemporal-unshared GC}
\hypertarget{sec:appendix-a}{}
The derivation from the feature updating process of DSTD-GC is presented in Eq. \ref{eq:stgc-dynamic}, Eq. \ref{eq:stgc-s-dynamic}, and Eq. \ref{eq:stgc-t-dynamic}, respectively. First, the feature updating of DS-GC for joint $q$ of frame $n$ is rewritten as:
\begin{equation}
    \mathbf{y}^{s(i)}_{qn} = \sum_{p}^{J} a^{s(i)}_{npq} \mathbf{x}^{(i)}_{pn} \boldsymbol{W}_{1},
    \label{eq:dsgc-single}
\end{equation}
where $a^{s(i)}_{npq}$ is from the spatial dynamic constrained correlation $A^{s(i)}$ in Eq. \ref{eq:dsgc-correlation}, and $i$ is added to highlight the sample-specific attribute. According to the spatiotemporal-equivalence between DS-GC and DT-GC, the feature updating of DT-GC can be formulated as:
\begin{equation}
    \mathbf{y}^{t(i)}_{qn} = \sum_{m}^{T} a^{t(i)}_{qmn} \mathbf{x}^{(i)}_{qm} \boldsymbol{W}_{2}.
    \label{eq:dtgc-single}
\end{equation}

By alternatively applying Eq. \ref{eq:dsgc-single} and \ref{eq:dtgc-single}, DSTD-GC can be formulated as:
\begin{equation}
\mathbf{y}^{st(i)}_{qn} = \sum_{p}^{J} \sum_{m}^{T} a^{s(i)}_{npq} a^{t(i)}_{qmn} \mathbf{x}^{(i)}_{pm} \boldsymbol{W}_1 \boldsymbol{W}_2.
\label{eq:dstdgc-single}
\end{equation}
It can be seen that Eq. \ref{eq:stgc-dynamic} is equivalent to Eq. \ref{eq:dstdgc-single} where $\boldsymbol{W} = \boldsymbol{W}_{1} \boldsymbol{W}_{2}$. Similarly, Eq. \ref{eq:dstdgc-single} can be reformulated as follows:
\begin{equation}
\mathbf{y}^{st(i)}_{qn} = \sum_{p}^{J} a^{s(i)}_{npq} \sum_{m}^{T} \mathbf{x}^{(i)}_{pm} (a^{t(i)}_{qmn} \boldsymbol{W}_1 \boldsymbol{W}_2),
\label{eq:dstdgc-s}
\end{equation}
\begin{equation}
\mathbf{y}^{st(i)}_{qn} = \sum_{m}^{T} a^{t(i)}_{qmn} \sum_{p}^{J} \mathbf{x}^{(i)}_{pm} (a^{s(i)}_{npq} \boldsymbol{W}_1 \boldsymbol{W}_2).
\label{eq:dstdgc-t}
\end{equation}
It can be seen that Eq. \ref{eq:stgc-s-dynamic} is equivalent to Eq. \ref{eq:dstdgc-s} and Eq. \ref{eq:stgc-t-dynamic}is equivalent to Eq. \ref{eq:dstdgc-t}.

\section*{B. Time Complexity of DSTD-GC}
\hypertarget{sec:appendix-b}{}
The time complexity analysis of STS-GC and DSTD-GC is presented here. As a GC consists of feature transformation and feature aggregation \cite{kipf2016semi}, we first show the time complexity of these two operations is the same for both GCs. Comparing Eq. \ref{eq:stgc-dynamic} with \ref{eq:stgc-static}, we find the time complexity is both $O(JTCC')$ for feature transformation and $O((J+T)JTC')$ for feature aggregation. For clarity, we utilize the frame-joint proximity assumption in Sect. \ref{sec:dstdgc-dis} and assume that $C = C'$. Under these assumptions, the overall time complexity of the basic graph convolution operations for both GCs is $O(CT^2(C+T))$.

In addition to basic graph convolution operations, DSTD-GC conducts constrained dynamic correlation modeling to extract dynamic correspondence adjustments. The modeling process does not increase the time complexity. Taking spatial modeling as an example, the most time-consuming operations are pair-wise concatenation and its following $MLP$, as illustrated in Fig. \ref{fig:details} (b). The time complexity for these two operations is $O(J^2T^2C/r)$. In our implementation, a very large $r$ is used, so $O(TC/r)$ can be approximated to $O(C)$. Thus, the overall time complexity of spatial constrained dynamic correlation modeling is $O(J^2TC)$. Combining the assumptions in the previous paragraph, the overall time complexity for the constrained dynamic correlation modeling is $O(CT^3)$, which is less or equal to the time complexity of the basic graph convolution operations. Therefore, the overall time complexity of DSTD-GC is $O(CT^2(C+T))$.

With the above analysis, it can be concluded that STS-GC and DSTD-GC have the same time complexity, and the extra operations in DSTD-GC don't increase the time complexity.

\bibliographystyle{IEEEtran}
\bibliography{main}{}

\begin{thebibliography}{10}
\providecommand{\url}[1]{#1}
\csname url@samestyle\endcsname
\providecommand{\newblock}{\relax}
\providecommand{\bibinfo}[2]{#2}
\providecommand{\BIBentrySTDinterwordspacing}{\spaceskip=0pt\relax}
\providecommand{\BIBentryALTinterwordstretchfactor}{4}
\providecommand{\BIBentryALTinterwordspacing}{\spaceskip=\fontdimen2\font plus
\BIBentryALTinterwordstretchfactor\fontdimen3\font minus
  \fontdimen4\font\relax}
\providecommand{\BIBforeignlanguage}[2]{{%
\expandafter\ifx\csname l@#1\endcsname\relax
\typeout{** WARNING: IEEEtran.bst: No hyphenation pattern has been}%
\typeout{** loaded for the language `#1'. Using the pattern for}%
\typeout{** the default language instead.}%
\else
\language=\csname l@#1\endcsname
\fi
#2}}
\providecommand{\BIBdecl}{\relax}
\BIBdecl

\bibitem{paden2016survey}
B.~Paden, M.~Čáp, S.~Z. Yong, D.~Yershov, and E.~Frazzoli, ``A survey of
  motion planning and control techniques for self-driving urban vehicles,''
  \emph{IEEE Transactions on Intelligent Vehicles}, vol.~1, no.~1, pp. 33--55,
  2016.

\bibitem{kong2018human}
Y.~Kong and Y.~Fu, ``Human action recognition and prediction: A survey,''
  \emph{arXiv preprint arXiv:1806.11230}, 2018.

\bibitem{unhelkar2018human}
V.~V. Unhelkar, P.~A. Lasota, Q.~Tyroller, R.-D. Buhai, L.~Marceau, B.~Deml,
  and J.~A. Shah, ``Human-aware robotic assistant for collaborative assembly:
  Integrating human motion prediction with planning in time,'' \emph{IEEE
  Robotics and Automation Letters}, vol.~3, no.~3, pp. 2394--2401, 2018.

\bibitem{yan2022probabilistic}
Z.~Yan, W.~He, Y.~Wang, L.~Sun, and X.~Yu, ``Probabilistic motion prediction
  and skill learning for human-to-cobot dual-arm handover control,'' \emph{IEEE
  Transactions on Neural Networks and Learning Systems}, pp. 1--13, 2022.

\bibitem{troje2002decomposing}
N.~F. Troje, ``Decomposing biological motion: A framework for analysis and
  synthesis of human gait patterns,'' \emph{Journal of Vision}, vol.~2, no.~5,
  pp. 2--2, 2002.

\bibitem{taylor2010dynamical}
G.~W. Taylor, L.~Sigal, D.~J. Fleet, and G.~E. Hinton, ``Dynamical binary
  latent variable models for 3d human pose tracking,'' in \emph{Proceedings of
  the IEEE Conference on Computer Vision and Pattern Recognition}, 2010, pp.
  631--638.

\bibitem{lehrmann2014efficient}
A.~M. Lehrmann, P.~V. Gehler, and S.~Nowozin, ``Efficient nonlinear markov
  models for human motion,'' in \emph{Proceedings of the IEEE Conference on
  Computer Vision and Pattern Recognition}, 2014, pp. 1314--1321.

\bibitem{li2018convolutional}
C.~Li, Z.~Zhang, W.~S. Lee, and G.~H. Lee, ``Convolutional sequence to sequence
  model for human dynamics,'' in \emph{Proceedings of the IEEE Conference on
  Computer Vision and Pattern Recognition}, 2018, pp. 5226--5234.

\bibitem{liu2020trajectorycnn}
X.~Liu, J.~Yin, J.~Liu, P.~Ding, J.~Liu, and H.~Liu, ``Trajectorycnn: A new
  spatio-temporal feature learning network for human motion prediction,''
  \emph{IEEE Transactions on Circuits and Systems for Video Technology},
  vol.~31, no.~6, pp. 2133--2146, 2021.

\bibitem{al2020attention}
A.~F. Al-aqel and M.~A. Khan, ``Attention mechanism for human motion
  prediction,'' in \emph{Proceedings of the International Conference on
  Computer Applications \& Information Security}, 2020, pp. 1--6.

\bibitem{guo2019human}
X.~Guo and J.~Choi, ``Human motion prediction via learning local structure
  representations and temporal dependencies,'' in \emph{Proceedings of the AAAI
  Conference on Artificial Intelligence}, 2019, pp. 2580--2587.

\bibitem{martinez2017human}
J.~Martinez, M.~J. Black, and J.~Romero, ``On human motion prediction using
  recurrent neural networks,'' in \emph{Proceedings of the IEEE Conference on
  Computer Vision and Pattern Recognition}, 2017, pp. 2891--2900.

\bibitem{fragkiadaki2015recurrent}
K.~Fragkiadaki, S.~Levine, P.~Felsen, and J.~Malik, ``Recurrent network models
  for human dynamics,'' in \emph{Proceedings of the IEEE International
  Conference on Computer Vision}, 2015, pp. 4346--4354.

\bibitem{tang2018long}
Y.~Tang, L.~Ma, W.~Liu, and W.~Zheng, ``Long-term human motion prediction by
  modeling motion context and enhancing motion dynamic,'' \emph{arXiv preprint
  arXiv:1805.02513}, 2018.

\bibitem{liu2019towards}
Z.~Liu, S.~Wu, S.~Jin, Q.~Liu, S.~Lu, R.~Zimmermann, and L.~Cheng, ``Towards
  natural and accurate future motion prediction of humans and animals,'' in
  \emph{Proceedings of the IEEE Conference on Computer Vision and Pattern
  Recognition}, 2019, pp. 10\,004--10\,012.

\bibitem{pavllo2020modeling}
D.~Pavllo, C.~Feichtenhofer, M.~Auli, and D.~Grangier, ``Modeling human motion
  with quaternion-based neural networks,'' \emph{International Journal of
  Computer Vision}, vol. 128, no.~4, pp. 855--872, 2020.

\bibitem{shu2021spatiotemporal}
X.~Shu, L.~Zhang, G.-J. Qi, W.~Liu, and J.~Tang, ``Spatiotemporal co-attention
  recurrent neural networks for human-skeleton motion prediction,'' \emph{IEEE
  Transactions on Pattern Analysis and Machine Intelligence}, vol.~44, no.~6,
  pp. 3300--3315, 2022.

\bibitem{wang2021pvred}
H.~Wang, J.~Dong, B.~Cheng, and J.~Feng, ``Pvred: A position-velocity recurrent
  encoder-decoder for human motion prediction,'' \emph{IEEE Transactions on
  Image Processing}, vol.~30, pp. 6096--6106, 2021.

\bibitem{gui2018adversarial}
L.-Y. Gui, Y.-X. Wang, X.~Liang, and J.~M. Moura, ``Adversarial geometry-aware
  human motion prediction,'' in \emph{Proceedings of the European Conference on
  Computer Vision}, 2018, pp. 786--803.

\bibitem{hernandez2019human}
A.~Hernandez, J.~Gall, and F.~Moreno-Noguer, ``Human motion prediction via
  spatio-temporal inpainting,'' in \emph{Proceedings of the IEEE International
  Conference on Computer Vision}, 2019, pp. 7134--7143.

\bibitem{cui2021efficient}
Q.~Cui, H.~Sun, Y.~Kong, X.~Zhang, and Y.~Li, ``Efficient human motion
  prediction using temporal convolutional generative adversarial network,''
  \emph{Information Sciences}, vol. 545, pp. 427--447, 2021.

\bibitem{ke2019learning}
Q.~Ke, M.~Bennamoun, H.~Rahmani, S.~An, F.~Sohel, and F.~Boussaid, ``Learning
  latent global network for skeleton-based action prediction,'' \emph{IEEE
  Transactions on Image Processing}, vol.~29, pp. 959--970, 2019.

\bibitem{kundu2019bihmp}
J.~N. Kundu, M.~Gor, and R.~V. Babu, ``Bihmp-gan: Bidirectional 3d human motion
  prediction gan,'' in \emph{Proceedings of the AAAI Conference on Artificial
  Intelligence}, 2019, pp. 8553--8560.

\bibitem{cai2020learning}
Y.~Cai, L.~Huang, Y.~Wang, T.-J. Cham, J.~Cai, J.~Yuan, J.~Liu, X.~Yang,
  Y.~Zhu, X.~Shen \emph{et~al.}, ``Learning progressive joint propagation for
  human motion prediction,'' in \emph{Proceedings of the European Conference on
  Computer Vision}, 2020, pp. 226--242.

\bibitem{aksan2021spatio}
E.~Aksan, M.~Kaufmann, P.~Cao, and O.~Hilliges, ``A spatio-temporal transformer
  for 3d human motion prediction,'' in \emph{Proceedings of the International
  Conference on 3D Vision}, 2021, pp. 565--574.

\bibitem{tevet2023human}
G.~Tevet, S.~Raab, B.~Gordon, Y.~Shafir, A.~H. Bermano, and D.~Cohen-or,
  ``Human motion diffusion model,'' in \emph{Proceedings of the International
  Conference on Learning Representations}, 2023.

\bibitem{mao2019learning}
W.~Mao, M.~Liu, M.~Salzmann, and H.~Li, ``Learning trajectory dependencies for
  human motion prediction,'' in \emph{Proceedings of the IEEE International
  Conference on Computer Vision}, 2019, pp. 9489--9497.

\bibitem{mao2020history}
W.~Mao, M.~Liu, and M.~Salzmann, ``History repeats itself: Human motion
  prediction via motion attention,'' in \emph{Proceedings of the European
  Conference on Computer Vision}, 2020, pp. 474--489.

\bibitem{li2020dynamic}
M.~Li, S.~Chen, Y.~Zhao, Y.~Zhang, Y.~Wang, and Q.~Tian, ``Dynamic multiscale
  graph neural networks for 3d skeleton based human motion prediction,'' in
  \emph{Proceedings of the IEEE Conference on Computer Vision and Pattern
  Recognition}, 2020, pp. 214--223.

\bibitem{cui2020learning}
Q.~Cui, H.~Sun, and F.~Yang, ``Learning dynamic relationships for 3d human
  motion prediction,'' in \emph{Proceedings of the IEEE conference on Computer
  Vision and Pattern Recognition}, 2020, pp. 6519--6527.

\bibitem{dang2021msr}
L.~Dang, Y.~Nie, C.~Long, Q.~Zhang, and G.~Li, ``Msr-gcn: Multi-scale residual
  graph convolution networks for human motion prediction,'' in
  \emph{Proceedings of the IEEE International Conference on Computer Vision},
  2021, pp. 11\,467--11\,476.

\bibitem{liu2021motion}
Z.~Liu, P.~Su, S.~Wu, X.~Shen, H.~Chen, Y.~Hao, and M.~Wang, ``Motion
  prediction using trajectory cues,'' in \emph{Proceedings of the IEEE
  International Conference on Computer Vision}, 2021, pp. 13\,299--13\,308.

\bibitem{sofianos2021space}
T.~Sofianos, A.~Sampieri, L.~Franco, and F.~Galasso, ``Space-time-separable
  graph convolutional network for pose forecasting,'' in \emph{Proceedings of
  the IEEE International Conference on Computer Vision}, 2021, pp.
  11\,209--11\,218.

\bibitem{li2022online}
M.~Li, S.~Chen, Y.~Shen, G.~Liu, I.~W. Tsang, and Y.~Zhang, ``Online
  multi-agent forecasting with interpretable collaborative graph neural
  networks,'' \emph{IEEE Transactions on Neural Networks and Learning Systems},
  pp. 1--15, 2022.

\bibitem{dong2022skeleton}
M.~Dong and C.~Xu, ``Skeleton-based human motion prediction with privileged
  supervision,'' \emph{IEEE Transactions on Neural Networks and Learning
  Systems}, pp. 1--14, 2022.

\bibitem{yan2018spatial}
S.~Yan, Y.~Xiong, and D.~Lin, ``Spatial temporal graph convolutional networks
  for skeleton-based action recognition,'' in \emph{Proceedings of the AAAI
  Conference on Artificial Intelligence}, 2018.

\bibitem{su2021motion}
P.~Su, Z.~Liu, S.~Wu, L.~Zhu, Y.~Yin, and X.~Shen, ``Motion prediction via
  joint dependency modeling in phase space,'' in \emph{Proceedings of the ACM
  International Conference on Multimedia}, 2021, pp. 713--721.

\bibitem{liu2021aggregated}
Z.~Liu, K.~Lyu, S.~Wu, H.~Chen, Y.~Hao, and S.~Ji, ``Aggregated multi-gans for
  controlled 3d human motion prediction,'' in \emph{Proceedings of the AAAI
  Conference on Artificial Intelligence}, 2021, pp. 2225--2232.

\bibitem{lyu2021learning}
K.~Lyu, Z.~Liu, S.~Wu, H.~Chen, X.~Zhang, and Y.~Yin, ``Learning human motion
  prediction via stochastic differential equations,'' in \emph{Proceedings of
  the ACM International Conference on Multimedia}, 2021, pp. 4976--4984.

\bibitem{derr2018signed}
T.~Derr, Y.~Ma, and J.~Tang, ``Signed graph convolutional networks,'' in
  \emph{Proceedings of the IEEE International Conference on Data Mining}, 2018,
  pp. 929--934.

\bibitem{qian2021pu}
G.~Qian, A.~Abualshour, G.~Li, A.~Thabet, and B.~Ghanem, ``Pu-gcn: Point cloud
  upsampling using graph convolutional networks,'' in \emph{Proceedings of the
  IEEE Conference on Computer Vision and Pattern Recognition}, 2021, pp.
  11\,683--11\,692.

\bibitem{yu2017spatio}
B.~Yu, H.~Yin, and Z.~Zhu, ``Spatio-temporal graph convolutional networks: A
  deep learning framework for traffic forecasting,'' \emph{arXiv preprint
  arXiv:1709.04875}, 2017.

\bibitem{li2021multiscale}
M.~Li, S.~Chen, Y.~Zhao, Y.~Zhang, Y.~Wang, and Q.~Tian, ``Multiscale
  spatio-temporal graph neural networks for 3d skeleton-based motion
  prediction,'' \emph{IEEE Transactions on Image Processing}, vol.~30, pp.
  7760--7775, 2021.

\bibitem{kipf2016semi}
T.~N. Kipf and M.~Welling, ``Semi-supervised classification with graph
  convolutional networks,'' \emph{arXiv preprint arXiv:1609.02907}, 2016.

\bibitem{ionescu2013human3}
C.~Ionescu, D.~Papava, V.~Olaru, and C.~Sminchisescu, ``Human3.6m: Large scale
  datasets and predictive methods for 3d human sensing in natural
  environments,'' \emph{IEEE Transactions on Pattern Analysis and Machine
  Intelligence}, vol.~36, no.~7, pp. 1325--1339, 2014.

\bibitem{marcard2018recovering}
T.~von Marcard, R.~Henschel, M.~J. Black, B.~Rosenhahn, and G.~Pons-Moll,
  ``Recovering accurate 3d human pose in the wild using imus and a moving
  camera,'' in \emph{Proceedings of the European Conference on Computer
  Vision}, 2018, pp. 601--617.

\bibitem{paszke2019pytorch}
A.~Paszke, S.~Gross, F.~Massa, A.~Lerer, J.~Bradbury, G.~Chanan, T.~Killeen,
  Z.~Lin, N.~Gimelshein, L.~Antiga \emph{et~al.}, ``Pytorch: An imperative
  style, high-performance deep learning library,'' in \emph{Proceedings of the
  Advances in Neural Information Processing Systems}, 2019, pp. 8024--8035.

\bibitem{he2015delving}
K.~He, X.~Zhang, S.~Ren, and J.~Sun, ``Delving deep into rectifiers: Surpassing
  human-level performance on imagenet classification,'' in \emph{Proceedings of
  the IEEE International Conference on Computer Vision}, 2015, pp. 1026--1034.

\bibitem{kingma2014adam}
D.~P. Kingma and J.~Ba, ``Adam: {A} method for stochastic optimization,'' in
  \emph{Proceedings of the International Conference on Learning
  Representations}, 2015.

\bibitem{mersch2021selfsupervised}
B.~Mersch, X.~Chen, J.~Behley, and C.~Stachniss, ``Self-supervised point cloud
  prediction using 3d spatio-temporal convolutional networks,'' \emph{ArXiv},
  vol. abs/2110.04076, 2021.

\bibitem{gomes2021spatiotemporal}
P.~S. Gomes, S.~Rossi, and L.~Toni, ``Spatio-temporal graph-rnn for point cloud
  prediction,'' in \emph{Proceedings of the IEEE International Conference on
  Image Processing}, 2021, pp. 3428--3432.

\bibitem{yang2016optimized}
H.-F. Yang, T.~S. Dillon, and Y.-P.~P. Chen, ``Optimized structure of the
  traffic flow forecasting model with a deep learning approach,'' \emph{IEEE
  Transactions on Neural Networks and Learning Systems}, vol.~28, no.~10, pp.
  2371--2381, 2016.

\bibitem{yang2021novel}
B.~Yang, G.~Yan, P.~Wang, C.-Y. Chan, X.~Song, and Y.~Chen, ``A novel
  graph-based trajectory predictor with pseudo-oracle,'' \emph{IEEE
  Transactions on Neural Networks and Learning Systems}, vol.~33, no.~12, pp.
  7064--7078, 2021.

\bibitem{shi2021sgcn}
L.~Shi, L.~Wang, C.~Long, S.~Zhou, M.~Zhou, Z.~Niu, and G.~Hua, ``Sgcn: Sparse
  graph convolution network for pedestrian trajectory prediction,'' in
  \emph{Proceedings of the IEEE Conference on Computer Vision and Pattern
  Recognition}, 2021, pp. 8994--9003.

\bibitem{kong2021spatiotemporal}
Y.~Kong, S.~Gao, Y.~Yue, Z.~Hou, H.~Shu, C.~Xie, Z.~Zhang, and Y.~Yuan,
  ``Spatio‐temporal graph convolutional network for diagnosis and treatment
  response prediction of major depressive disorder from functional
  connectivity,'' \emph{Human Brain Mapping}, vol.~42, pp. 3922--3933, 2021.

\bibitem{yu2021deepsg}
L.~Yu, B.~Du, X.~Hu, L.~Sun, L.~Han, and W.~Lv, ``Deep spatio-temporal graph
  convolutional network for traffic accident prediction,''
  \emph{Neurocomputing}, vol. 423, pp. 135--147, 2021.

\bibitem{liu2021light}
N.~Liu, W.~Zhao, D.~Zhang, J.~Han, and L.~Shao, ``Light field saliency
  detection with dual local graph learning and reciprocative guidance,'' in
  \emph{Proceedings of the IEEE International Conference on Computer Vision},
  2021, pp. 4712--4721.

\bibitem{liu2021instance}
N.~Liu, L.~Li, W.~Zhao, J.~Han, and L.~Shao, ``Instance-level relative saliency
  ranking with graph reasoning,'' \emph{IEEE Transactions on Pattern Analysis
  and Machine Intelligence}, vol.~44, no.~11, pp. 8321--8337, 2021.

\bibitem{zhang2020weakly}
D.~Zhang, W.~Zeng, J.~Yao, and J.~Han, ``Weakly supervised object detection
  using proposal- and semantic-level relationships,'' \emph{IEEE Transactions
  on Pattern Analysis and Machine Intelligence}, vol.~44, no.~6, pp.
  3349--3363, 2022.

\bibitem{liu2020comprehensive}
K.~Liu, R.~Ding, Z.~Zou, L.~Wang, and W.~Tang, ``A comprehensive study of
  weight sharing in graph networks for 3d human pose estimation,'' in
  \emph{Proceedings of the European Conference on Computer Vision}, 2020, pp.
  318--334.

\end{thebibliography}

\vfill
\end{document}